\pdfoutput=1
\documentclass{article}

\PassOptionsToPackage{numbers,compress}{natbib}
\usepackage[main, final]{neurips_2026}
\makeatletter
\renewcommand{\@noticestring}{Accepted at the 40th Conference on Neural Information Processing Systems (NeurIPS 2026); to appear in its proceedings.}
\makeatother

\usepackage[utf8]{inputenc}
\usepackage[T1]{fontenc}
\usepackage{hyperref}
\usepackage{url}
\usepackage{booktabs}
\usepackage{amsfonts}
\usepackage{amsmath}
\usepackage{amssymb}
\usepackage{amsthm}
\usepackage{nicefrac}
\usepackage{microtype}
\usepackage{xcolor}
\usepackage{algorithm}
\usepackage{algpseudocode}
\usepackage{graphicx}
\usepackage{subcaption}
\usepackage{multirow}

\theoremstyle{definition}
\newtheorem{definition}{Definition}[section]
\newtheorem{assumption}{Assumption}
\theoremstyle{plain}

\newtheorem{theorem}[definition]{Theorem}
\newtheorem{corollary}[definition]{Corollary}

\title{STABLE: A Continual Learning Optimizer with Adaptive Drift Control}
\author{%
  Xiaoyan Li \\
  Department of Computer Science \\
  University of Pisa \\[0.5ex]
  Department of AI, Data and Decision Sciences \\
  Luiss Guido Carli University \\
  \texttt{x.li7@studenti.unipi.it} \\
  \And
  Lanpei Li \\
  Institute of Information Science \\
  and Technologies ``Alessandro Faedo'' (ISTI) \\
  National Research Council of Italy (CNR) \\[0.5ex]
  Department of Computer Science \\
  University of Pisa \\
  \texttt{lanpei.li@isti.cnr.it} \\
  \AND
  Massimo Coppola \\
  Institute of Information Science \\
  and Technologies ``Alessandro Faedo'' (ISTI) \\
  National Research Council of Italy (CNR) \\
  \texttt{massimo.coppola@isti.cnr.it} \\
  \And
  Vincenzo Lomonaco \\
  Department of AI, Data and Decision Sciences \\
  Luiss Guido Carli University \\
  \texttt{vlomonaco@luiss.it} \\
  \AND
  Antonio Carta \\
  Department of Computer Science \\
  University of Pisa \\
  \texttt{antonio.carta@unipi.it} \\
}

\begin{document}

\maketitle

\begin{abstract}
Continual learning suffers from catastrophic forgetting, leading to the loss of past knowledge and unstable adaptation across task streams. Existing approaches typically address this issue through replay buffers, architectural changes, or additional regularization terms. We argue that forgetting is a consequence of unstable learning dynamics, which may benefit from an optimizer-level solution. The main challenge is that learning dynamics are time-dependent, as demonstrated by empirical phenomena such as the stability gap.
First, we formally prove that, for distance-based classifiers, bounded parameter drift guarantees bounded margin erosion under a local score-stability condition. Based on this insight, we propose STABLE, a primal–dual optimizer with adaptive drift control. STABLE enforces a drift budget by increasing the dual variable when the budget is violated and decreasing it otherwise. This results in high stability without penalizing plasticity. Experiments on long-horizon unimodal streams and CLIP-based multimodal streams show that vanilla PEFT baselines trained with STABLE match or outperform specialized continual learning methods. In the multimodal setting, STABLE further preserves zero-shot transfer, matching or exceeding dedicated multimodal continual learning methods. Code is available at \url{https://github.com/atlas-luiss/STABLE}.
\end{abstract}
\section{Introduction}

Continual learning (CL) studies how to adapt a model to a stream of tasks while retaining previously acquired knowledge~\cite{Bell2025FutureCL,Quarantiello2026WinterCL}. Most CL methods mitigate forgetting by adding task-specific machinery: replay revisits stored examples from earlier tasks~\cite{Buzzega2020}, regularization penalizes changes to important parameters~\cite{Kirkpatrick2017} or distills previous predictions~\cite{Li2016}, and prompt-, routing-, or adapter-based methods isolate capacity across tasks~\cite{Wang2022,Wang2022dual,Liang2024}. These approaches can be effective, but they increase training complexity and sometimes inference cost. Yet they leave a basic question underexplored: \emph{when, and how, is forgetting introduced during training?}

Recent continual-evaluation studies show that performance on previous tasks can drop sharply during the early transient that follows a task switch, revealing a task-boundary stability gap~\cite{DeLange2023StabilityGap}. This suggests that a substantial fraction of forgetting may occur during this short post-switch transient: at task $t{+}1$, optimization is driven by a new objective and can rapidly move away from parameters that supported previous predictions.

Learning instabilities have also been observed in multimodal models such as CLIP, where each task switch must preserve not only class separation but also image--text alignment~\cite{Yu2026MMCLSurvey}. Existing CLIP-based continual learning methods therefore often add dedicated multimodal mechanisms, including modality-gap compensation~\cite{Huang2025ICCV}, adaptive representation adjustment and parameter fusion~\cite{Huang2024RAPF}, mixture-of-experts adapters~\cite{Yu2024MoE4CL}, probabilistic finetuning~\cite{Jha2024CLAP4CLIP}, and external knowledge injection~\cite{Zhou2025ENGINE}.

Standard optimizers can also contribute to forgetting. For example, adaptive optimizers such as Adam and AdamW are particularly fragile in CL settings. When the optimizer state is carried across task boundaries, new updates are scaled according to the geometry of the previous task, causing an abrupt departure from the previous-task solution before optimization on the new task has stabilized.

Overall, we argue that these phenomena arise from unstable learning dynamics and optimizer-level issues. While architectural and replay-based CL solutions can mitigate forgetting, they struggle to adapt to the time-dependent dynamics of forgetting. In this paper, we propose an optimizer-level solution in which stability constraints are time-dependent and adapted to the current learning dynamics.

First, we formally prove that, for distance-based classifiers, margin erosion can be bounded via parameter drift. Then, we propose \textbf{STABLE}, a primal–dual optimizer that automatically adapts the stability–plasticity tradeoff according to the learning dynamics. Given a fixed drift budget, STABLE increases the dual variable when the budget is violated, strengthening the pullback toward the inherited solution. When drift falls below the budget, it decreases the dual variable to recover plasticity. As a result, STABLE addresses the stability gap by automatically increasing the pullback strength after a task switch. In multimodal settings, STABLE preserves image–text alignment through trajectory stabilization rather than modality-specific continual learning modules.

We evaluate STABLE with two standard parameter-efficient architectures, LoRA and prompts, in both unimodal vision and multimodal settings. Across these settings, simply replacing Adam with STABLE substantially improves continual retention and makes vanilla PEFT baselines competitive with methods that rely on dedicated continual learning machinery. Together, these results support the view that, in the settings we evaluate, task-boundary instability is a major source of forgetting and that controlling post-switch drift can improve continual learning in both unimodal and multimodal models.

\textbf{Contributions.}
\begin{enumerate}
    \item We formalize prediction stability under parameter drift for distance-based classifiers, and show that, under local score stability, inherited margins certify prediction-preserving neighborhoods.
    \item We introduce STABLE, a primal--dual optimizer that enforces a soft drift budget around the inherited solution, strengthening the pullback during the early post-switch transient and relaxing it once the drift returns within the budget.
    \item With pre-trained ViT and CLIP backbones, replacing Adam with STABLE lets vanilla LoRA and prompt tuning match or outperform specialized continual learning methods, and keeps CLIP's zero-shot transfer close to its pre-trained level.
\end{enumerate}

\section{Prediction Stability Under Parameter Drift}
\label{sec:motivation}

In this section, we define forgetting for distance-based classifiers in terms of margin erosion, i.e., the loss of predictive margin caused by representation drift. We prove that margin erosion can be bounded by parameter drift. This result provides a simple measure of forgetting that is easy to compute and strongly correlated with empirical forgetting (Spearman $0.94$--$0.97$, Section~\ref{sec:local_score_margin}). It also motivates the use of distance-based classifiers, which have been shown to outperform softmax classifiers empirically~\cite{Mensink2012Metric}.

\subsection{Setup and Notation}
Let $\theta$ denote the trainable parameters updated during continual adaptation, and let $\theta_t^\star$ denote their values at the end of task $t$, which we call the inherited solution for subsequent tasks. Let $\mathcal Y_{\le t}$ denote the set of all classes seen up to task $t$. For an input $x$, the task-$t$ head produces a score vector over $\mathcal Y_{\le t}$, denoted by $s_t(x;\theta)\in\mathbb R^{|\mathcal Y_{\le t}|}$. For each class $c \in \mathcal Y_{\le t}$, we write $s_{t,c}(x;\theta) = [s_t(x;\theta)]_c$. The task-$t$ prediction is the class with the largest score.

We consider inference heads where class scores are induced directly by the current representation map(s), with $h_\theta : \mathcal X \to \mathbb R^d$ denoting the input representation map:
\begin{equation}
s_{t,c}(x;\theta)=g_t\bigl(h_\theta(x),\,r_{t,c}[\mathcal H_\theta]\bigr),
\label{eq:head_form}
\end{equation}
where $\mathcal H_\theta$ denotes the representation map(s) used by the head, $r_{t,c}[\mathcal H_\theta]$ is a representation-based descriptor of class $c$, and $g_t$ is a scoring rule. For example, in a prototype-based classifier, $r_{t,c}$ is the centroid of class $c$ in representation space and $g_t$ is cosine similarity. In CLIP zero-shot classification, $h_\theta(x)$ is the normalized image embedding, $r_{t,c}$ is the normalized text embedding of a class prompt, and $g_t$ is the scaled dot product. The key structural property is that both the input representation and the class descriptors are determined by representation maps, rather than by independently updated classifier weights. This makes parameter drift translate directly into changes in the old-task score vector, without the confounding effect of independently trained classifier parameters.

\subsection{Margin Erosion Under Parameter Drift}

For a labeled example $(x,y)$ with $y\in\mathcal Y_{\le t}$, define the task-$t$ margin at parameters $\theta$ as
\[
\gamma_t(x,y;\theta) = s_{t,y}(x;\theta) - \max_{\substack{c \in \mathcal Y_{\le t} \\ c \ne y}} s_{t,c}(x;\theta).
\]
At the inherited solution $\theta_t^\star$, a positive margin ensures correct prediction. After subsequent adaptation, the prediction can change only if this gap shrinks sufficiently for a competing class to overtake the true class.

We refer to the decrease in margin caused by moving from $\theta_t^\star$ to $\theta$ as margin erosion:
\begin{equation}
\label{eq:margin_erosion_def}
\Delta\gamma_t(x,y;\theta) = \gamma_t(x,y;\theta_t^\star)-\gamma_t(x,y;\theta).
\end{equation}
Positive values indicate margin shrinkage; negative values indicate margin improvement.

We impose a local stability condition on the task-$t$ score vector around the inherited solution. This condition requires each class score to change at most linearly with the distance moved within a small neighborhood of $\theta_t^\star$.
\begin{assumption}[Local score stability]
\label{ass:score_stability}
There exist a neighborhood $\mathcal U_t$ of $\theta_t^\star$ and a constant $C_s > 0$ such that, for every input $x$ and every $\theta \in \mathcal U_t$,
\[
\|s_t(x;\theta) - s_t(x;\theta_t^\star)\|_\infty \le C_s\|\theta - \theta_t^\star\|.
\]
\end{assumption}
This is a local Lipschitz condition on the score function, which is weaker than
a representation-level Lipschitz condition. Prediction changes are determined
by score differences, and many changes in the representation may leave the
induced class scores nearly unchanged. Section~\ref{sec:local_score_margin}
empirically evaluates this stability within the drift-controlled neighborhood
produced by STABLE and estimates the corresponding local sensitivity from
observed score changes.

Under Assumption~\ref{ass:score_stability}, the erosion of any old-task margin is bounded by how far the parameters have moved.
\begin{theorem}[Margin erosion under bounded parameter drift]
\label{thm:margin_erosion}
Suppose Assumption~\ref{ass:score_stability} holds, with neighborhood $\mathcal U_t$ and constant $C_s > 0$. Then, for any labeled example $(x,y)$ with $y\in\mathcal Y_{\le t}$ and any $\theta\in\mathcal U_t$,
\[
\Delta \gamma_t(x,y;\theta)\le 2C_s\|\theta-\theta_t^\star\|.
\]
\end{theorem}

\textbf{Proof sketch.}
By Assumption~\ref{ass:score_stability}, each class score changes by at most $C_s\|\theta-\theta_t^\star\|$. The true-class score can decrease by at most this amount, while every competing-class score can increase by at most the same amount, so the margin can shrink by at most $2C_s\|\theta-\theta_t^\star\|$. The full proof is given in Appendix~\ref{app:margin_erosion_proof}.

\section{Stabilizing Training Adaptation with Budgeted Lagrangian Enforcement (STABLE)}
\label{sec:method}
In this section, we introduce Stabilizing Training Adaptation with Budgeted Lagrangian Enforcement (STABLE), an optimizer for continual learning.

Regularization methods such as EWC~\cite{Kirkpatrick2017}, SI~\cite{Zenke2017SI}, MAS~\cite{Aljundi2018MAS}, and L2-SP~\cite{Li2018L2SP} penalize parameter drift with a coefficient fixed before training. EWC, SI, and MAS also weight each parameter by its importance. STABLE instead applies one coefficient to all parameters and treats it as a time-dependent control variable. This is important because continual learning exhibits time-dependent training dynamics: the stability gap~\cite{DeLange2023StabilityGap} shows that forgetting can occur during the early post-switch transient, while later adaptation may require weaker constraints to preserve plasticity. Thus, a static coefficient imposes one stability--plasticity tradeoff throughout training, whereas STABLE adapts the effective coefficient through $\lambda_k$ to enforce a fixed drift budget. The adapted coefficient also differs across tasks: Figure~\ref{fig:post_switch_controller} shows different levels at Tasks~10, 50, and~100. In experiments with the same LoRA model, STABLE outperforms both EWC and an L2 pullback with a tuned fixed coefficient~(Appendices~\ref{app:combined} and~\ref{app:fixed_pullback}).

Section~\ref{sec:motivation} established that limiting drift from the inherited solution bounds old-task margin erosion. We therefore constrain how far the parameters move from the previous-task solution during training.

Let $\rho_t(\theta) = \|\theta - \theta_t^\star\|$ denote the distance from the inherited solution $\theta_t^\star$, which is obtained at the end of task $t$ and used during training on task $t+1$. Let $\tau > 0$ denote the drift budget, which acts as a safety radius: if Assumption~\ref{ass:score_stability} holds on the ball of radius $\tau$ around $\theta_t^\star$, Theorem~\ref{thm:margin_erosion} bounds old-task margin erosion inside it by $2C_s\tau$. A smaller $\tau$ thus constrains forgetting more tightly but leaves less room to fit the current task; a larger $\tau$ allows more adaptation at the cost of weaker retention.
We therefore formulate training on task $t+1$ as minimizing the current-task loss under the drift budget $\tau$:
\[
\min_\theta\ \mathcal{L}_{t+1}(\theta) \quad \text{s.t.} \quad \rho_t(\theta) \le \tau,
\]
where $\mathcal L_{t+1}$ denotes the current-task loss.
The associated Lagrangian is
\[
\mathcal{J}(\theta,\lambda) = \mathcal{L}_{t+1}(\theta) + \lambda\bigl(\rho_t(\theta)^2 - \tau^2\bigr), \quad \lambda \ge 0.
\]
Using $\rho_t(\theta)^2$ rather than $\rho_t(\theta)$ avoids the discontinuous subgradient of the L2 norm at the anchor and instead yields the smooth gradient $2\lambda(\theta - \theta_t^\star)$. We solve the Lagrangian with a soft primal--dual update rather than a hard projection (Appendix~\ref{app:hard_projection} compares against the corresponding hard-projection baseline): each iteration alternates a primal Adam~\cite{Kingma2015} update of $\theta$ with a dual update of $\lambda$.

At iteration $k$, the primal update forms the augmented gradient
\[
\tilde g_k = \nabla_\theta \mathcal{L}_{t+1}(\theta_k) + 2\lambda_k(\theta_k - \theta_t^\star).
\]
It then applies the standard Adam update with this gradient. The second term acts as a soft pullback toward the inherited solution, with an effective strength controlled by $\lambda_k$ rather than a fixed coefficient.

After the primal step, we compute the post-update distance $\rho_{k+1} = \|\theta_{k+1} - \theta_t^\star\|$ and update the dual variable according to
\begin{equation}
\lambda_{k+1} \leftarrow
\begin{cases}
\lambda_k + \eta\bigl((\rho_{k+1}^2-\tau_+^2)+\bar r_k\bigr),
& \rho_{k+1} > \tau_+, \\[5pt]
\lambda_k \left(\dfrac{\rho_{k+1}^2}{\tau_-^2}\right)^\eta,
& \rho_{k+1} < \tau_-, \\[8pt]
\lambda_k,
& \text{otherwise},
\end{cases}
\label{eq:duco_dual_update}
\end{equation}
where $\eta>0$ is the dual step size. The term $\bar r_k=\frac{1}{k-1}\sum_{i=1}^{k-1}[\rho_i^2-\tau^2]_+$ is the running average of the constraint violation over previous steps within the current task, and $\tau_\pm=\tau(1\pm\epsilon_{\mathrm{dz}})$ with $\epsilon_{\mathrm{dz}}=0.05$ defines a deadzone around the constraint boundary.

When the distance exceeds $\tau_+$, $\lambda$ increases, strengthening the pullback toward $\theta_t^\star$. The $\bar r_k$ term makes the controller more responsive when the trajectory repeatedly overshoots the budget during the post-switch transient. When the distance falls below $\tau_-$, $\lambda$ is multiplicatively damped. This allows the strong pullback accumulated during the boundary transient to decay quickly, restoring plasticity for the current task once the trajectory returns to the safe region. $\lambda$ remains fixed within the deadzone to avoid oscillation. Because $\lambda$ adapts the pullback to each task, a single budget, calibrated once from Task~1, can be used for the whole stream (Appendix~\ref{app:calibration}).

Algorithm~\ref{alg:stable} in Appendix~\ref{app:algorithm} summarizes the full procedure.

\section{Experiments}
\label{sec:experiments}

\subsection{Experimental Setup}
\textbf{Datasets and task streams.} We evaluate in two settings that stress task-boundary instability in complementary ways. In unimodal vision, we use long-horizon class-incremental streams on ImageNet-R~\cite{Hendrycks2021ImageNetR} (100 tasks, 99 task switches), CIFAR-100~\cite{Krizhevsky2009CIFAR} (50 tasks, 49 task switches), and ImageNet-A-100~\cite{Hendrycks2021ImageNetA} (the 100 most frequent ImageNet-A classes; 50 tasks, 49 task switches), with two classes per task. The dense sequence of task boundaries repeatedly exposes the optimizer to post-switch transients. For multimodal vision-language experiments, we use 10-task streams on the same datasets with 20, 10, and 10 classes per task. This setting tests whether adaptation can preserve both stream discrimination and the pre-trained image-text alignment required for zero-shot transfer.

\textbf{Backbones, PEFT modules, and training.} We freeze pre-trained ViT-B/16~\cite{Dosovitskiy2021ViT} backbones (ImageNet-21K for vision; CLIP~\cite{Radford2021CLIP} for multimodal) and restrict adaptation to parameter-efficient modules. Full-model finetuning and training from scratch are therefore outside our scope (Appendix~\ref{app:limitations}). LoRA~\cite{Hu2022} (rank $r=8$, scaling $\alpha=16$) is injected into the self-attention projections: packed QKV and output projections in the unimodal ViT implementation, and Q/K/V/O projections in both CLIP towers. For prompt tuning, we use key/value prefix prompts of length 20 per transformer block, without prompt pools, routing, or retrieval. All vanilla PEFT runs use a learning rate of $5\times10^{-4}$, weight decay $10^{-4}$, and Adam coefficients $(0.9, 0.999)$. Adam and STABLE share the architecture, data stream, and these hyperparameters; besides the drift constraint, they differ only in the batch size of some unimodal streams (Appendix~\ref{app:experimental_details}). We train each task for 40, 20, and 50 epochs on unimodal ImageNet-R, CIFAR-100, and ImageNet-A-100, and for 20 epochs in the multimodal setting. For STABLE, the dual step size is fixed to $\eta=0.03$; accuracy varies little with $\eta$, staying within 1.4 points across a 60-fold range on 100-task ImageNet-R (Appendix~\ref{app:eta_sensitivity}). The initial dual value is $\lambda_{\mathrm{init}}=1.0$, and the drift budget $\tau$ is calibrated from Task~1 (Appendix~\ref{app:calibration}).

\textbf{Evaluation protocol and metrics.}
For task-agnostic inference, we use a covariance-whitened cosine nearest-class-mean (W-NCM) head, following prototype-based inference in class-incremental learning~\cite{Mensink2012Metric,Rebuffi2017iCaRL} and covariance-aware extensions for anisotropic feature spaces~\cite{Goswami2023FeCAM}. This head updates class prototypes and a shared covariance estimate in a single pass over each current-task dataset; at inference, it whitens features and prototypes, then predicts by cosine similarity. Because the prototypes and whitening matrix are both functions of the current representation map $h_\theta$, W-NCM instantiates the representation-induced head form of Section~\ref{sec:motivation}: parameter drift translates directly into score change, with no independently updated classifier to confound the source of forgetting. We use W-NCM for all unimodal experiments; it also gives the tested baselines higher accuracy than their native linear heads (Appendix~\ref{app:linear_head}). In multimodal, we use W-NCM for in-stream classification of vanilla LoRA runs, and the standard CLIP similarity head for zero-shot transfer. For specialized multimodal methods, we report their original inference procedures in the main table, since inference design is integral to those methods; a common W-NCM evaluation for all methods is in Appendix Table~\ref{tab:multimodal_shared_wncm_full}. We report final average accuracy and average forgetting following standard practice~\cite{DeLange2023StabilityGap,Smith2023CODAPrompt}.

\textbf{Baselines.}
We compare vanilla PEFT trained with Adam vs.\ STABLE with the settings above. We also compare against specialized CL methods using the authors' official implementations and original training protocols. In unimodal vision: O-LoRA~\cite{Wang2023OLoRA}, HiDeLoRA~\cite{Wang2025HiDePET}, InfLoRA~\cite{Liang2024} (LoRA-based); L2P~\cite{Wang2022}, DualPrompt~\cite{Wang2022dual}, CODA-Prompt~\cite{Smith2023CODAPrompt} (prompt-based). In multimodal: MindTheGap~\cite{Huang2025ICCV}, RAPF~\cite{Huang2024RAPF}, MoE4CL~\cite{Yu2024MoE4CL}, CLAP4CLIP~\cite{Jha2024CLAP4CLIP}, and ENGINE~\cite{Zhou2025ENGINE}.

Appendix~\ref{app:experimental_details} gives dataset construction, training settings, and metric definitions.

\subsection{Long-Horizon Vision Class-Incremental Learning}
\label{sec:unimodal_results}
We first evaluate STABLE in long-horizon unimodal class-incremental learning, where the 100-task and 50-task streams repeatedly stress post-switch stability. Table~\ref{tab:unimodal_unified} compares vanilla PEFT architectures trained using Adam or STABLE against specialized continual learning baselines from the same architectural family.

\begin{table}[t]
\caption{Unimodal class-incremental results. 3-seed means; std in Appendix Table~\ref{tab:full_results}. \textbf{Bold}: best, \underline{underlined}: second.}
\label{tab:unimodal_unified}
\centering
\small
{\setlength{\tabcolsep}{3pt}
\resizebox{\textwidth}{!}{%
\begin{tabular}{@{}llccccccccc@{}}
\toprule
Method & Opt. & \multicolumn{3}{c}{ImageNet-R} & \multicolumn{3}{c}{CIFAR-100} & \multicolumn{3}{c}{ImageNet-A-100} \\
\cmidrule(lr){3-5} \cmidrule(lr){6-8} \cmidrule(l){9-11}
& & Params & $\bar{A}_T$ (\%) & $\bar{F}_T$ (\%) & Params & $\bar{A}_T$ (\%) & $\bar{F}_T$ (\%) & Params & $\bar{A}_T$ (\%) & $\bar{F}_T$ (\%) \\
\midrule
Joint LoRA & Adam & 0.44M & 79.85 & -- & 0.44M & 92.32 & -- & 0.44M & 72.68 & -- \\
Joint Prompts & Adam & 0.37M & 72.23 & -- & 0.37M & 88.86 & -- & 0.37M & 61.34 & -- \\
\midrule
\multicolumn{11}{l}{\textit{LoRA-based methods}} \\
O-LoRA~\cite{Wang2023OLoRA} & AdamW & 44.2M & 57.87 & 33.87 & 22.1M & 74.09 & 25.90 & 22.1M & 26.92 & 59.37 \\
HiDeLoRA~\cite{Wang2025HiDePET} & Adam & 10.04M & 72.36 & \underline{6.16} & 6.20M & 79.92 & \textbf{2.29} & 6.20M & 60.37 & 10.62 \\
InfLoRA~\cite{Liang2024} & Adam & 36.86M & 72.54 & 18.05 & 18.43M & 84.93 & 13.74 & 18.43M & \underline{61.69} & 22.33 \\
Vanilla LoRA & Adam & 0.44M & 0.66 & 60.22 & 0.44M & 1.30 & 75.70 & 0.44M & 1.25 & 58.41 \\
Vanilla LoRA (ours) & \textbf{STABLE} & 0.44M & \textbf{73.12} & 10.17 & 0.44M & \textbf{88.77} & 7.90 & 0.44M & \textbf{62.34} & 12.11 \\
\midrule
\multicolumn{11}{l}{\textit{Prompt-based methods}} \\
L2P~\cite{Wang2022} & Adam & 0.94M & 68.60 & 8.74 & 0.94M & 85.32 & 8.31 & 0.94M & 19.13 & \underline{9.07} \\
DualPrompt~\cite{Wang2022dual} & Adam & 0.95M & 70.39 & 10.69 & 0.95M & 86.25 & 9.21 & 0.95M & 19.37 & \textbf{7.62} \\
CODA-Prompt~\cite{Smith2023CODAPrompt} & Adam & 3.84M & 69.32 & \textbf{6.07} & 3.84M & 87.72 & 5.35 & 3.84M & 59.37 & 10.67 \\
Vanilla Prompts & Adam & 0.37M & 68.01 & 16.15 & 0.37M & 83.88 & 12.00 & 0.37M & 56.31 & 22.13 \\
Vanilla Prompts (ours) & \textbf{STABLE} & 0.37M & \underline{72.99} & 7.64 & 0.37M & \underline{87.88} & \underline{4.99} & 0.37M & 58.88 & 14.23 \\
\bottomrule
\end{tabular}
}
}
\end{table}

Table~\ref{tab:unimodal_unified} shows a consistent pattern. Under Adam, vanilla LoRA collapses on all three streams. It has no mechanism that protects the previous tasks' solution; such unprotected finetuning is known to forget severely in class-incremental learning~\cite{Masana2023Survey}, and Adam has been observed to perform worse as more tasks arrive~\cite{Elsayed2024UPGD}. The collapse comes from the many task switches: with 10 tasks, vanilla LoRA with Adam still reaches 64.34\% on multimodal ImageNet-R (Table~\ref{tab:multimodal_unified}), but with 50 or 100 tasks the unconstrained drift after every switch (Figure~\ref{fig:post_switch_controller}) erodes old-task margins until their predictions flip. Replacing Adam with STABLE makes the same LoRA architecture competitive with specialized LoRA baselines while using far fewer trainable parameters and maintaining low forgetting. Vanilla prompts trained with STABLE similarly match or exceed specialized prompt methods in final average accuracy while maintaining competitive forgetting scores. Overall, STABLE makes standard PEFT architectures competitive with specialized continual learning methods without adding task-specific modules, prompt routing, or replay. This suggests that much of the benefit often sought through specialized architectures can also be obtained by stabilizing the optimizer trajectory at task boundaries.

\begin{table}[b]
\caption{CLIP multimodal results (10 tasks). 3-seed means; std in Appendix Table~\ref{tab:multimodal_unified_full}. \textbf{Bold}: best, \underline{underlined}: second. CLAP4CLIP uses replay; all other methods are replay-free.}
\label{tab:multimodal_unified}
\centering
\small
{\setlength{\tabcolsep}{2pt}
\renewcommand{\arraystretch}{1.12}
\begin{tabular}{@{}llccccccc@{}}
\toprule
Method & Opt. & Params & \multicolumn{2}{c}{ImageNet-R} & \multicolumn{2}{c}{CIFAR-100} & \multicolumn{2}{c}{ImageNet-A-100} \\
\cmidrule(lr){4-5} \cmidrule(lr){6-7} \cmidrule(l){8-9}
& & & $\bar{A}_T$ (\%) & $\bar{F}_T$ (\%) & $\bar{A}_T$ (\%) & $\bar{F}_T$ (\%) & $\bar{A}_T$ (\%) & $\bar{F}_T$ (\%) \\
\midrule
Joint LoRA & Adam & 0.98M & 86.06 & -- & 87.45 & -- & 74.28 & -- \\
\midrule
Vanilla LoRA & Adam & 0.98M & 64.34 & 25.74 & 56.84 & 44.68 & 28.15 & 55.03 \\
\midrule
MindTheGap~\cite{Huang2025ICCV} & Adam & 0.49M & \textbf{81.50} & 6.94 & 80.40 & 9.62 & 65.98 & 6.59 \\
RAPF~\cite{Huang2024RAPF} & Adam & 0.26M & 80.41 & \underline{3.30} & \underline{80.41} & \textbf{6.99} & 61.07 & 16.57 \\
MoE4CL~\cite{Yu2024MoE4CL} & AdamW & 4.03M & 80.37 & 5.48 & 79.08 & 9.64 & 58.20 & \textbf{5.75} \\
CLAP4CLIP~\cite{Jha2024CLAP4CLIP} & SGD & 9.46M & 79.37 & 7.90 & 78.61 & 9.92 & \underline{67.27} & 11.69 \\
ENGINE~\cite{Zhou2025ENGINE} & SGD & 5.25M & 76.80 & 4.86 & 73.85 & 12.62 & 57.22 & 11.52 \\
Vanilla LoRA (ours) & \textbf{STABLE} & 0.98M & \underline{81.28} & \textbf{2.80} & \textbf{81.61} & \underline{9.10} & \textbf{70.19} & \underline{6.54} \\
\bottomrule
\end{tabular}
}
\end{table}

\subsection{Multimodal Continual Learning with CLIP}
\label{sec:multimodal_results}
We next evaluate STABLE in multimodal continual learning with CLIP ViT-B/16 on 10-task ImageNet-R, CIFAR-100, and ImageNet-A-100 streams. This setting requires the model to retain stream discrimination while preserving CLIP’s pre-trained image-text alignment. Because LoRA is more sensitive to task-boundary instability than prompt tuning, we focus on LoRA and evaluate both in-stream classification and zero-shot transfer.

\textbf{A. In-stream classification.}

Table~\ref{tab:multimodal_unified} reports in-stream results. STABLE reduces forgetting substantially compared to Adam (e.g., from 25.74\% to 2.80\% on ImageNet-R) and improves accuracy on all three streams, with the largest margin on ImageNet-A-100, where it raises accuracy from 28.15\% to 70.19\%.

\textbf{B. Zero-shot transfer retention.}

In-stream performance alone does not capture whether continual adaptation preserves CLIP's open-vocabulary recognition capability. Since zero-shot classification depends on the alignment between image and text embeddings, any degradation of this alignment directly harms transfer to unseen classes. We therefore evaluate zero-shot transfer after the final stream task on a held-out suite of datasets: Food101~\cite{Bossard2014Food101}, Oxford-IIIT Pets~\cite{Parkhi2012Pets}, Flowers102~\cite{Nilsback2008Flowers}, DTD~\cite{Cimpoi2014DTD}, and Stanford Cars~\cite{Krause2013Cars}.

\begin{table}[t]
\caption{CLIP zero-shot accuracy (\%) after ImageNet-R/CIFAR-100 streams. 3-seed means ($\pm$ std and ImageNet-A-100 in Appendix Tables~\ref{tab:multimodal_zeroshot_full_imagenet_r}--\ref{tab:multimodal_zeroshot_full_imagenet_a}); \textbf{Bold}: best, \underline{underlined}: second.}
\label{tab:multimodal_zeroshot}
\centering
\small
{\setlength{\tabcolsep}{1.2pt}
\renewcommand{\arraystretch}{1.15}
\resizebox{\linewidth}{!}{
\begin{tabular}{@{}lcccccccccc@{}}
\toprule
Method & \multicolumn{5}{c}{ImageNet-R} & \multicolumn{5}{c}{CIFAR-100} \\
\cmidrule(lr){2-6} \cmidrule(l){7-11}
& Food101 & Pets & Flowers & DTD & Cars & Food101 & Pets & Flowers & DTD & Cars \\
\midrule
Pre-trained CLIP & 88.38 & 88.01 & 66.03 & 44.31 & 64.42 & 88.38 & 88.01 & 66.03 & 44.31 & 64.42 \\
\midrule
Joint LoRA + Adam & 88.32 & 89.76 & 68.35 & 44.75 & 63.10 & 86.35 & 88.16 & 68.68 & 44.29 & 61.61 \\
Vanilla LoRA + Adam & 74.76 & 75.80 & 52.51 & 38.67 & 48.90 & 58.68 & 62.66 & 33.85 & 32.94 & 38.20 \\
\midrule
MindTheGap & 85.97 & \underline{87.40} & \underline{65.43} & \underline{43.78} & \underline{60.32} & \underline{87.85} & \underline{86.21} & \textbf{68.81} & \underline{44.72} & \underline{61.73} \\
RAPF & 47.01 & 69.91 & 32.06 & 24.77 & 23.90 & 18.71 & 26.02 & 10.73 & 15.05 & 6.17 \\
MoE4CL & 85.12 & 84.45 & 61.66 & 43.62 & 59.69 & 86.35 & 80.43 & 66.88 & 41.35 & 55.48 \\
CLAP4CLIP & \underline{86.20} & 80.34 & 58.91 & 40.60 & 59.51 & 85.94 & 80.65 & 60.25 & 41.22 & 60.17 \\
ENGINE & 37.57 & 59.05 & 22.41 & 20.94 & 17.54 & 20.46 & 23.26 & 16.98 & 13.12 & 9.15 \\
Vanilla LoRA + STABLE (ours) & \textbf{88.56} & \textbf{88.77} & \textbf{67.86} & \textbf{46.05} & \textbf{64.19} & \textbf{88.63} & \textbf{88.62} & \underline{67.73} & \textbf{44.84} & \textbf{64.11} \\
\bottomrule
\end{tabular}
}
}
\end{table}

Table~\ref{tab:multimodal_zeroshot} reports zero-shot transfer. STABLE stays close to the pre-trained CLIP baseline across all five datasets on both streams, achieving the highest transfer scores among continual methods on 9 of 10 evaluations, and improves over Adam everywhere. Under Adam, transfer degrades sharply --- on CIFAR-100, Food101 accuracy drops from 88.38\% to 58.68\%. Appendix Table~\ref{tab:multimodal_zeroshot_full_imagenet_a} shows the same recovery pattern after ImageNet-A-100.

These results show that task-boundary instability degrades both in-stream performance and the pre-trained image-text alignment required for zero-shot recognition. By constraining post-switch drift, STABLE preserves both in-stream performance and image-text alignment without adding multimodal-specific continual learning modules.

\section{Boundary Dynamics}
\label{sec:analysis}

Section~\ref{sec:experiments} shows that replacing Adam with STABLE improves retention and transfer. We next test whether these gains come from controlling post-switch drift during adaptation to a new task. Specifically, we ask whether (i) the dual variable responds to boundary-induced drift as intended, (ii) old-task scores and margins remain locally stable after pullback toward the inherited solution, and (iii) margin erosion aligns with observed forgetting.

\subsection{Dual Response to Post-Switch Drift}
\label{sec:post_switch_controller}

We inspect the optimization trajectory after representative task switches on the ImageNet-R 100-task LoRA run, probing at Tasks~10, 50, and~100 (early, middle, and late in the stream). At each boundary, STABLE stores the inherited parameters as the anchor $\theta_t^{\star}$ and monitors $\|\theta-\theta_t^{\star}\|$ and $\lambda$. For comparison, we also plot the corresponding Adam distance, measured from Adam's own inherited solution.

\begin{figure}[b]
    \centering{\includegraphics[trim={5 7 10 10},clip,width=0.9\linewidth]{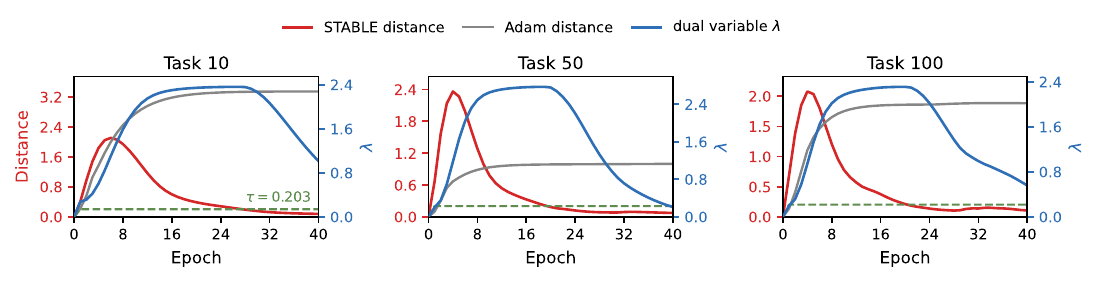}}
    \caption{STABLE drift-control dynamics on ImageNet-R 100-task LoRA. STABLE distance from $\theta_t^\star$ (red), Adam distance (gray). When drift exceeds the budget $\tau$ (green dashed), $\lambda$ (blue) rises; once the trajectory returns, $\lambda$ decays.}
    \label{fig:post_switch_controller}
\end{figure}

Figure~\ref{fig:post_switch_controller} shows a consistent pattern. During the early epochs after each task switch, the distance from $\theta_t^\star$ increases sharply. This rise reflects fast learning of the new task. When drift exceeds the budget, $\lambda$ rises and strengthens the pullback; the trajectory is drawn back, after which $\lambda$ decays. Adam's drift grows without constraint. The dual response acts as a soft boundary guard rather than a hard projection: it reacts strongly when the post-switch trajectory moves too far, but relaxes once the transient is contained. This limits the duration of the high-risk boundary phase without permanently suppressing plasticity on the new task, because every primal step, even during the pullback, still includes the current-task gradient.

Over the whole stream, the growth of forgetting slows down. In this run, average forgetting rises only from 8.63\% at Task~50 to 9.88\% at Task~100, and the oldest tasks are forgotten no more than the middle ones. We attribute this to the drift control described above and to the anchor moving at each boundary to a solution that already supports all previous tasks (Appendix~\ref{app:long_stream}).

Resetting Adam's moments at task boundaries does not prevent collapse (0.77\% accuracy, 89.87\% forgetting; Appendix~\ref{app:adam_reset}). Static drift-control baselines also underperform STABLE: a fixed L2 pullback remains below STABLE even after coefficient tuning, and a hard-projection baseline remains below STABLE across projection budgets (Appendices~\ref{app:fixed_pullback} and~\ref{app:hard_projection}). Neither simple fix is sufficient, confirming that drift control must adapt online.

\subsection{Local Margin Stability and Boundary Forgetting}
\label{sec:local_score_margin}

Theorem~\ref{thm:margin_erosion} bounds old-task margin erosion in terms of parameter drift around the inherited solution. We therefore examine the portion of each post-switch trajectory that lies inside the drift budget. For each probed boundary $t\in\{10,50,100\}$, we collect training steps satisfying $\|\theta_i-\theta^\star_{t-1}\|\le\tau$. At these steps, we measure worst-case projected margin erosion relative to the inherited solution, clipping margin improvements to zero so that gains on some examples do not mask losses on others. For each boundary, we estimate the empirical local sensitivity constant $\hat C_s$ from end-of-epoch calibration probes and evaluate the corresponding margin-erosion bound $2\hat C_s\|\theta-\theta_{t-1}^\star\|$ on held-out intermediate probes. Appendix~\ref{app:local_score_stability} provides the full diagnostic. We then examine the correlation between projected margin erosion and old-task forgetting.

\begin{figure}[t]
    \centering{\includegraphics[trim={5 10 5 18},clip,width=\linewidth]{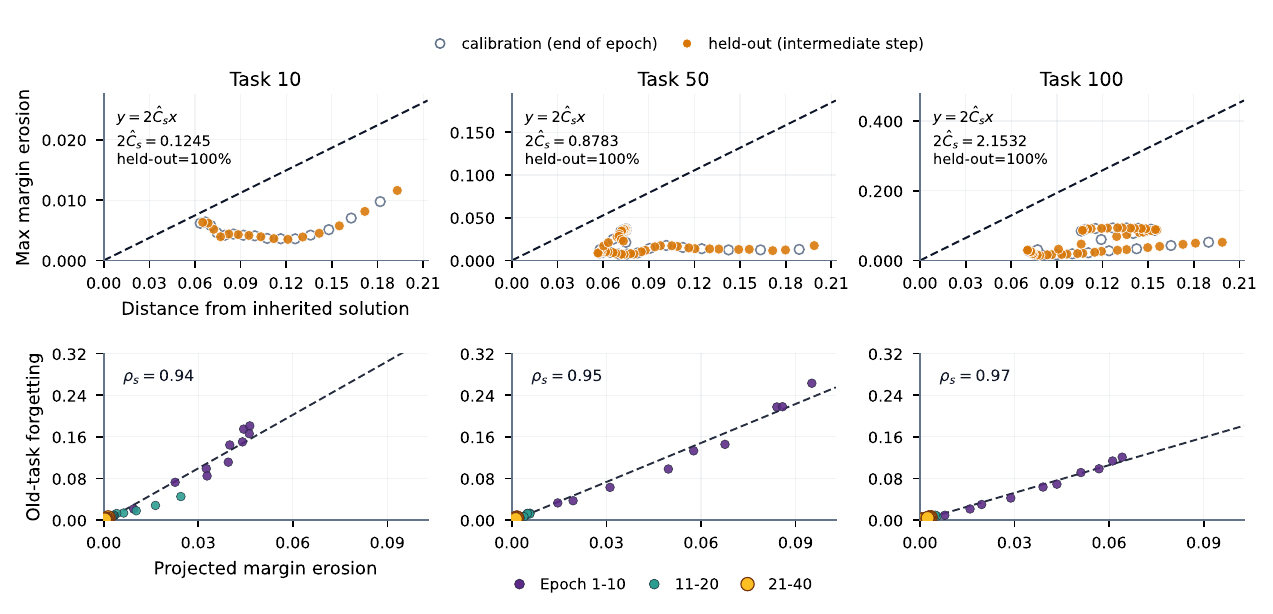}}
    \caption{Local margin stability and boundary forgetting on ImageNet-R 100-task LoRA. Top: worst-case projected margin erosion vs.~distance from $\theta_{t-1}^\star$ inside the drift budget (dashed: empirical bound from end-of-epoch probes). Bottom: projected margin erosion vs.~old-task forgetting (one point per post-switch epoch).}
    \label{fig:local_margin_forgetting}
\end{figure}

The top row of Figure~\ref{fig:local_margin_forgetting} shows that, once the trajectory lies inside the drift budget, worst-case projected margin erosion remains below the corresponding empirical bound. The bottom row shows a strong monotone relationship between projected margin erosion and old-task forgetting, with Spearman correlations $\rho_s=0.94$, $0.95$, and $0.97$ for Tasks~10, 50, and~100. Epochs with little margin erosion exhibit little forgetting, while epochs with larger erosion show larger drops in old-task accuracy. This connects the local margin analysis to the retention gains in Section~\ref{sec:experiments}: STABLE improves retention by limiting the post-switch drift that erodes inherited margins.

\subsection{Zero-Shot Transfer Margins in CLIP}
\label{sec:clip_zs_retention}

The margin analysis also applies to CLIP zero-shot transfer. Here, continual adaptation must preserve not only stream discrimination but also the pre-trained separation between image and text embeddings for unseen classes. We track zero-shot accuracy and margins after each ImageNet-R task, using the standard CLIP head on Food101 and Oxford-IIIT Pets. Task~0 denotes pre-trained CLIP; Tasks~1--10 correspond to the model after each stream task. The zero-shot margin $\gamma_{\mathrm{zs}}(x,y;\theta)$ is the similarity of the correct class prompt minus the largest competing similarity.

\begin{figure}[t]
    \centering{\includegraphics[trim={5 8 5 10},clip,width=0.8\linewidth]
    {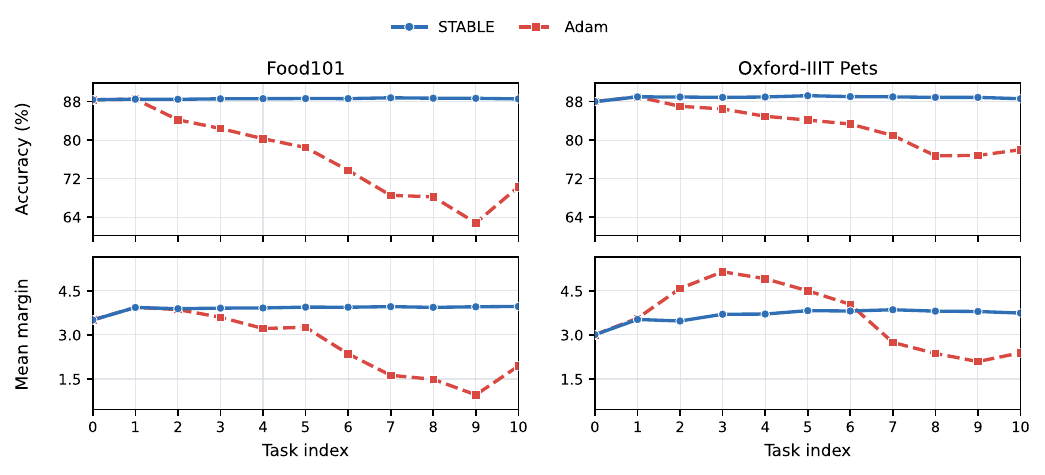}}
    \caption{Zero-shot transfer retention on ImageNet-R 10-task CLIP LoRA. Task~0: pre-trained CLIP; Tasks~1--10: stream tasks. Top: accuracy; bottom: mean margin.}
    \label{fig:clip_zs_task_retention}
\end{figure}

Figure~\ref{fig:clip_zs_task_retention} shows that Adam progressively reduces both zero-shot accuracy and image-text margins, especially on Food101. STABLE keeps both close to pre-trained levels throughout the stream.

\section{Related Work}
\label{sec:related}

\textbf{Replay, regularization, and PEFT-based methods.}
Most CL methods mitigate forgetting by modifying the data stream, training objective, or model architecture. Replay methods interleave stored examples from previous tasks with current-task data~\cite{Buzzega2020}. Regularization methods penalize changes to parameters or predictions important for earlier tasks, as in EWC~\cite{Kirkpatrick2017} and LwF~\cite{Li2016}. PEFT-based methods adapt pre-trained models through parameter-efficient modules~\cite{Coleman2026PECFT}: prompt methods learn prompt tokens with task-conditioned routing or mixing~\cite{Wang2022,Wang2022dual,Smith2023CODAPrompt}, while LoRA-based methods reduce interference through task-structured or subspace-constrained low-rank updates~\cite{Wang2023OLoRA,Liang2024,Wang2025HiDePET}. These approaches add stability through auxiliary data, objectives, or capacity allocation. STABLE instead changes the optimizer, leaving the data stream, loss, and PEFT architecture unchanged.

\textbf{Incremental learning in dynamic environments.}
Yu et al.~\cite{Yu2026GILCD} formalize streams whose input distribution and label space evolve together, and adapt to them with training-free prototype calibration. Chen and Zhou~\cite{Chen2025DisCo} observe that introducing domain shift across class-incremental tasks reduces forgetting, and promote the feature separation behind this effect with a prototype pool. In both works, drift is a property of the data stream, and the methods reshape the representation to handle it. STABLE instead addresses parameter drift, the movement of the parameters away from the inherited solution, and controls it in the optimizer.

\textbf{Task-boundary instability.}
CL performance can degrade sharply during the early post-switch transient, producing a stability gap between offline and continual learning~\cite{DeLange2023StabilityGap,soutif2023improving}. Related work links forgetting to abrupt representation changes under distribution shift~\cite{Caccia2022}, identifying the task boundary as a fragile phase of adaptation. We focus on this phase from an optimization perspective: post-switch drift from the inherited solution erodes old-task margins, so STABLE ties its controller to inherited-solution drift rather than to generic signals such as gradient norm or a preset penalty schedule.

\textbf{Optimization-based CL.}
Several continual learning methods reduce forgetting by using constrained optimization or orthogonal projection. GEM~\cite{LopezPaz2017GEM} uses gradients computed from episodic memory to constrain each update so that losses on previous tasks do not increase. OGD~\cite{Farajtabar2020OGD} projects new-task gradients onto directions that preserve previous-task predictions, while OWM~\cite{Zeng2019OWM} restricts weight updates to directions orthogonal to the input subspace of previous tasks, and OCAR~\cite{urettini2025online} uses the Fisher information. These methods protect past tasks through constraints derived from stored examples, gradients, or input subspaces. STABLE is motivated by a different view: forgetting arises from unstable, time-dependent training dynamics. Rather than projecting updates or imposing a fixed auxiliary penalty, STABLE controls drift from the inherited solution with a soft dual variable that responds to the current trajectory.

\textbf{CLIP-based multimodal CL.}
CLIP-based continual learning must preserve both stream discrimination and pre-trained image-text alignment~\cite{Yu2026MMCLSurvey}. Existing methods address this with multimodal-specific mechanisms, including modality-gap compensation~\cite{Huang2025ICCV}, adaptive representation fusion~\cite{Huang2024RAPF}, mixture-of-experts adapters~\cite{Yu2024MoE4CL}, probabilistic finetuning~\cite{Jha2024CLAP4CLIP}, and external knowledge injection~\cite{Zhou2025ENGINE}. %
STABLE shows that targeting optimization instability can mitigate the modality gap, preserving both in-stream accuracy and zero-shot transfer without adding CLIP-specific modules.

\section{Conclusion}
\label{sec:conclusion}

This work studies forgetting as a consequence of unstable optimization. %
We model forgetting as old-task margin erosion and prove that bounded parameter drift bounds it under local score stability. Empirically, margin erosion closely tracks forgetting.

Optimizer-level issues such as the stability gap can cause severe forgetting during the early post-switch transient.
STABLE addresses this failure mode by controlling drift from the inherited solution: its primal--dual update strengthens the pullback when drift exceeds the budget and relaxes it once training stabilizes. Across long-horizon vision and CLIP-based multimodal streams, replacing Adam with STABLE improves retention and zero-shot transfer with standard LoRA or prompt tuning, showing that stabilizing learning dynamics can reduce forgetting without specialized architectures.

\begin{ack}
This research was partially supported by the FIS2 Grant from the Italian Ministry of University and Research (MUR), Grant No.~FIS2023-03382, under the project ``Continual, Decentralized Compositionality for Sustainable Artificial Intelligence'' with Prof.~Vincenzo Lomonaco serving as Principal Investigator (PI). AC acknowledges support for the CoEvolution project, funded by EU Horizon 2020 under GA n 101168559.
We acknowledge ISCRA for awarding this project access to the LEONARDO supercomputer, owned by the EuroHPC Joint Undertaking, hosted by CINECA (Italy).
There are no competing interests to declare.
\end{ack}

\bibliographystyle{plainnat}
\bibliography{references}

\newpage
\appendix
\makeatletter
\setcounter{topnumber}{3}
\setcounter{totalnumber}{4}
\setlength{\@fptop}{0pt}
\setlength{\@fpsep}{\floatsep}
\makeatother

\section{Limitations}
\label{app:limitations}
We evaluate STABLE in the standard setting of this literature: parameter-efficient continual learning with pre-trained ViT and CLIP backbones on vision and vision-language streams. Our prediction-preservation guarantee and margin-erosion analysis focus on representation-induced distance-based heads, including W-NCM and CLIP image--text similarity heads, where parameter drift directly induces score drift. Although the optimizer is not architecturally tied to these heads, other heads and objectives are outside the scope of our current theoretical guarantee and empirical study. For independently parameterized classifier heads, a possible extension is to maintain separate drift budgets for shared representation parameters and newly introduced classifier weights, so that representation drift can be controlled while class-incremental head expansion remains flexible. Extending the analysis to such heads, as well as to generative objectives, full-model finetuning, training from scratch, other modalities, and confidence calibration~\cite{Li2024CalibrationCL}, remains future work.

The drift budget is calibrated once from Task~1, using only information available before later tasks arrive. The same calibration works on streams whose tasks differ in difficulty and training-set size (Appendix~\ref{app:calibration}); testing it with deliberately atypical first tasks is a natural next step.

Because STABLE acts only on the optimizer, it can also be applied to the trainable parameters of specialized PEFT methods such as InfLoRA~\cite{Liang2024}, HiDeLoRA~\cite{Wang2025HiDePET}, or CODA-Prompt~\cite{Smith2023CODAPrompt}. Our experiments combine it with vanilla LoRA and prompts and with EWC, LwF, and replay (Appendix~\ref{app:combined}); evaluating it with these specialized methods is future work.

\section{Algorithm Details}
\label{app:algorithm}

Algorithm~\ref{alg:stable} summarizes STABLE. In practice, Task~1 is used to collect update statistics for budget calibration, while drift control is activated from Task~2 onward. Because the loop indexes the current task, training on task $t>1$ uses the completed previous-task parameters $\theta_{t-1}^\star$ as the inherited anchor; this is the same constraint as Section~\ref{sec:method} with the index shifted from task $t{+}1$ to task $t$.

The anchor is one copy of the trainable PEFT parameters; it contains neither the frozen backbone nor any data from previous tasks. It is overwritten at every task boundary, so its memory cost stays constant over the stream (0.44M parameters for unimodal LoRA). The dual variable, by contrast, is carried over from one task to the next. Each task starts at the anchor, so the drift is zero at the switch and builds up during the early post-switch transient. Once it exceeds $\tau_+$, Eq.~\eqref{eq:duco_dual_update} raises $\lambda$ and strengthens the pullback, which draws the drift back within the budget, whereas Adam's drift levels off at 5--16 times the budget (Figure~\ref{fig:post_switch_controller}).

\begin{algorithm}[ht]
\caption{STABLE with Task-1 calibration}
\label{alg:stable}
\begin{algorithmic}[1]
\Require Current trainable parameters $\theta$, initial dual value $\lambda_{\mathrm{init}}=1.0$, primal learning rate $\alpha$, dual step size $\eta$, deadzone $\epsilon_{\mathrm{dz}}=0.05$, Adam coefficients $(\beta_1,\beta_2)$
\State Initialize persistent moments, dual variable, violation memory, and global step counter
\Statex \hspace{\algorithmicindent}$m \leftarrow 0$, $v \leftarrow 0$, $\lambda \leftarrow 0$, $\bar r \leftarrow 0$, $q \leftarrow 0$, $s \leftarrow 0$
\For{task $t = 1, 2, \ldots, T$}
    \State Reset task violation memory $\bar r \leftarrow 0$, $q \leftarrow 0$
    \If{$t = 1$}
        \State Disable drift control and record update magnitudes for calibration
    \Else
        \State Use inherited anchor $\theta_{t-1}^\star$ from the completed previous task
        \If{$t = 2$}
            \State Initialize $\lambda \leftarrow \lambda_{\mathrm{init}}$
        \EndIf
        \State Activate drift control using anchor $\theta_{t-1}^\star$ and drift budget $\tau$
    \EndIf
    \For{each minibatch of task $t$}
        \State $s \leftarrow s + 1$
        \State Compute current-task gradient
        \Statex \hspace{\algorithmicindent}\hspace{\algorithmicindent}$g \leftarrow \nabla_\theta \mathcal{L}_t(\theta)$
        \If{drift control is active}
            \State Form the augmented gradient
            \Statex \hspace{\algorithmicindent}\hspace{\algorithmicindent}\hspace{\algorithmicindent}$\tilde g \leftarrow g + 2\lambda(\theta - \theta_{t-1}^\star)$
        \Else
            \State Set $\tilde g \leftarrow g$
        \EndIf
        \State Update Adam moments
        \Statex \hspace{\algorithmicindent}\hspace{\algorithmicindent}$m \leftarrow \beta_1 m + (1-\beta_1)\tilde g$, $v \leftarrow \beta_2 v + (1-\beta_2)\tilde g^2$
        \State Form bias-corrected moments
        \Statex \hspace{\algorithmicindent}\hspace{\algorithmicindent}$\hat m \leftarrow m / (1-\beta_1^s)$, $\hat v \leftarrow v / (1-\beta_2^s)$
        \State Compute the primal step and update the parameters
        \Statex \hspace{\algorithmicindent}\hspace{\algorithmicindent}$\Delta\theta \leftarrow -\alpha\hat m / (\sqrt{\hat v}+\epsilon)$, $\theta \leftarrow \theta + \Delta\theta$
        \If{drift control is active}
            \State Compute the post-update distance
            \Statex \hspace{\algorithmicindent}\hspace{\algorithmicindent}\hspace{\algorithmicindent}$\rho \leftarrow \|\theta - \theta_{t-1}^\star\|_2$
            \State Update $\lambda$ using Eq.~\eqref{eq:duco_dual_update} with current $\bar r$
            \State $q \leftarrow q+1$;\quad $\bar r \leftarrow \frac{(q-1)\bar r + [\rho^2-\tau^2]_+}{q}$
        \Else
            \State Record $\|\Delta\theta\|_2^2$ for calibration
        \EndIf
    \EndFor
    \If{$t = 1$}
        \State Finalize the calibrated drift budget $\tau$ from the recorded Task-1 update statistics
    \EndIf
    \State Store completed-task parameters as anchor $\theta_t^\star \leftarrow \theta$
\EndFor
\end{algorithmic}
\end{algorithm}

\noindent Algorithm~\ref{alg:wncm} summarizes the W-NCM update and prediction procedure. The head maintains class prototypes $\{\mu_c\}$, a global mean $\mu$, and a shared covariance estimate $\Sigma$ accumulated over all seen tasks. After completing each task $t$, we update these statistics in a single pass over the task dataset $\mathcal D_t$ using Welford's algorithm; we apply diagonal shrinkage $(\rho_{\mathrm{sh}} = 0.07)$ for numerical stability. Given the feature extractor $\phi(\cdot)$ (the LoRA-adapted backbone), predictions use cosine similarity in the whitened space:
\[
\hat{y} = \arg\max_c \cos\!\left(\Sigma^{-1/2}\bigl(\phi(x)-\mu\bigr), \Sigma^{-1/2}\bigl(\mu_c-\mu\bigr)\right).
\]

\begin{algorithm}[ht]
\caption{W-NCM Update and Prediction}
\label{alg:wncm}
\small
\begin{algorithmic}[1]
\State \textbf{Hyper:} shrinkage $\rho_{\mathrm{sh}}\leftarrow 0.07$, eigen floor $\epsilon\leftarrow 10^{-12}$, norm floor $\epsilon_n\leftarrow 10^{-8}$.
\State \textbf{State:} global mean $\mu\in\mathbb R^d$, scatter $M\in\mathbb R^{d\times d}$, covariance $\Sigma\in\mathbb R^{d\times d}$, count $n$;
\State \hspace{\algorithmicindent}per-class mean $\{\mu_c\in\mathbb R^d\}$ and count $\{n_c\}$ for an ordered class list $\mathcal C$;
\State \hspace{\algorithmicindent}cached whitening matrix $W$ (initially \texttt{None}).
\State \textbf{Procedure} \textsc{Update}$(Z\in\mathbb R^{N\times d},\, y\in\mathbb N^N)$ $\{\text{Replay-free: update using current-task features only}\}$
\If{$n=0$}
    \State $\mu\leftarrow \frac{1}{N}\sum_{i=1}^{N} z_i$;\quad $M\leftarrow (Z-\mu)^\top(Z-\mu)$;\quad $n\leftarrow N$
\Else
    \State $n_{\mathrm{old}}\leftarrow n$;\quad $\mu_{\mathrm{old}}\leftarrow \mu$;\quad $n\leftarrow n_{\mathrm{old}}+N$
    \State $\mu\leftarrow \frac{n_{\mathrm{old}}\mu_{\mathrm{old}}+\sum_{i=1}^{N} z_i}{n}$ $\{\text{batched Welford mean}\}$
    \State $M\leftarrow M+(Z-\mu_{\mathrm{old}})^\top(Z-\mu)$ $\{\text{batched Welford scatter}\}$
\EndIf
\State $\Sigma\leftarrow
\begin{cases}
M/(n-1), & n>1\\
I, & n\le 1
\end{cases}$
\State $\Sigma\leftarrow (1-\rho_{\mathrm{sh}})\Sigma+\rho_{\mathrm{sh}}\operatorname{diag}(\Sigma)$ $\{\text{diagonal shrinkage}\}$
\State $W\leftarrow \texttt{None}$ $\{\text{invalidate cached whitening}\}$
\For{each class $c\in\operatorname{unique}(y)$}
    \State $m\leftarrow |\{i:y_i=c\}|$;\quad $s\leftarrow \sum_{i:y_i=c} z_i$
    \If{$c\notin\mathcal C$}
        \State append $c$ to $\mathcal C$;\quad $\mu_c\leftarrow s/m$;\quad $n_c\leftarrow m$
    \Else
        \State $\mu_c\leftarrow \frac{n_c\mu_c+s}{n_c+m}$;\quad $n_c\leftarrow n_c+m$
    \EndIf
\EndFor
\State \textbf{End Procedure}
\State \textbf{Procedure} \textsc{Predict}$(Z\in\mathbb R^{N\times d})$
\If{$|\mathcal C|=0$}
    \State \Return $0\in\mathbb N^N$ $\{\text{no prototypes yet}\}$
\EndIf
\If{$W=\texttt{None}$}
    \State eigendecompose $\Sigma=U\operatorname{diag}(\sigma)U^\top$
    \State $\sigma\leftarrow \max(\sigma,\epsilon)$
    \State $W\leftarrow U\operatorname{diag}(\sigma^{-1/2})U^\top$ $\{\Sigma^{-1/2}\}$
\EndIf
\State $\widetilde Z\leftarrow (Z-\mu)W$ $\{\text{whitened features}\}$
\State $P\leftarrow [\mu_c]_{c\in\mathcal C}\in\mathbb R^{|\mathcal C|\times d}$
\State $\widetilde P\leftarrow (P-\mu)W$ $\{\text{whitened prototypes}\}$
\State row-normalize: $\widetilde Z_i\leftarrow \widetilde Z_i/(\|\widetilde Z_i\|_2+\epsilon_n)$;\quad $\widetilde P_j\leftarrow \widetilde P_j/(\|\widetilde P_j\|_2+\epsilon_n)$
\State scores $S\leftarrow \widetilde Z\widetilde P^\top$ $\{\text{cosine similarity in whitened space}\}$
\State $k_i\leftarrow \arg\max_j S_{i,j}$;\quad $\hat y_i\leftarrow \mathcal C[k_i]$
\State \Return $\hat y$
\State \textbf{End Procedure}
\end{algorithmic}
\end{algorithm}

\section{Budget Calibration}
\label{app:calibration}

During Task~1, STABLE disables drift control and records squared update norms $u_k=\|\Delta\theta_k\|_2^2$ from the trainable parameter block. Let $T_1$ be the number of Task-1 optimizer steps. We use the last $N=\min\{32,\max(\lceil 0.1T_1\rceil,12)\}$ records, form rolling 5-step sums $s_j=\sum_{\ell=0}^{4}u_{j+\ell}$, and set the squared drift budget by the 75th percentile:
\[
\tau^2 = \frac{L}{5} Q_{75}(\{s_j\}), \qquad
L=\min\left\{\left\lceil\sqrt{\frac{1}{(1-\beta_1)(1-\beta_2)}}\right\rceil,256\right\}.
\]
With $(\beta_1,\beta_2)=(0.9,0.999)$, this gives $L=100$. This makes the controller adapt to the dataset-specific update scale while avoiding per-dataset manual tuning.

The calibrated budget is an estimate of the optimizer's typical drift range. It sets the safety radius of Section~\ref{sec:method} and need not be precise, because the self-adjusting $\lambda$ carries the flexibility across tasks. When a later task is very different, $\lambda$ rises and strengthens the pullback, and it decreases again once the drift is back within the budget. The penalty is also soft, so the model can exceed the budget when the current task needs it, at a price set by $\lambda$. ImageNet-A-100 tests the calibration in practice: some of its classes have only 26 images while others have several times more, so its tasks differ in both difficulty and training-set size (Appendix~\ref{app:experimental_details}). With the budget calibrated from its first task, vanilla LoRA with STABLE matches or outperforms the specialized methods on this stream (Table~\ref{tab:unimodal_unified}). In the multimodal experiments on the same dataset, accuracy is robust to this value: across seeds, $\tau^2$ ranges from $10^{-4}$ to $2.6\times10^{-3}$, while final accuracy has a standard deviation of 0.37 points (Table~\ref{tab:multimodal_unified_full}).

\section{Experimental Details}
\label{app:experimental_details}

\textbf{Datasets and task streams.}
In unimodal vision, we use long-horizon class-incremental streams on ImageNet-R, CIFAR-100, and ImageNet-A-100, partitioned into 100, 50, and 50 tasks, respectively, with 2 classes per task. In multimodal vision-language learning, we use 10-task streams on ImageNet-R, CIFAR-100 and ImageNet-A-100, with 20, 10 and 10 classes per task, respectively. For each seed, all compared methods use the same class order.

For ImageNet-A, we use the 100 most frequent classes from the public ImageNet-A dataset, which we denote as ImageNet-A-100. This avoids tail classes with only 2--5 images, which are unsuitable for a 50-task class-incremental stream with two classes per task. The selected subset ensures at least 26 images per class. We split each class into 80\% training and 20\% testing, yielding 4,839 training and 1,164 test images.

\textbf{Backbones and PEFT modules.}
In unimodal vision, we use an ImageNet-21K pre-trained ViT-B/16. In the multimodal setting, we use CLIP ViT-B/16. LoRA uses rank $r=8$ and scaling $\alpha=16$, and is applied to all self-attention projections in the unimodal model and to the self-attention layers of both the vision and text towers in CLIP, while keeping the projection layers and logit scale fixed. For prompting, we use key/value prefix prompts of length 20 inserted into every transformer block, without prompt pools, routing, or retrieval.

\textbf{Training protocol.}
In unimodal vision, vanilla PEFT runs with Adam use batch size 128; runs with STABLE use batch size 24 on ImageNet-A-100, for LoRA on ImageNet-R, and for prompts on CIFAR-100, and 128 otherwise. In the multimodal setting, all vanilla PEFT runs use batch size 128, except on ImageNet-A-100, which uses batch size 24 because of its smaller training set. All vanilla PEFT runs use learning rate $5\times10^{-4}$, weight decay $10^{-4}$, and Adam coefficients $\beta_1=0.9$ and $\beta_2=0.999$. In unimodal vision, each task is trained for 40 epochs on ImageNet-R, 20 on CIFAR-100, and 50 on ImageNet-A-100, without a learning-rate schedule. In the multimodal setting, each task is trained for 20 epochs using a flat-then-cosine schedule, with the learning rate held constant for the first 60\% of training and then decayed to $0.01\times$ its initial value. For STABLE, the dual step size $\eta$ is fixed to 0.03 and the drift budget $\tau$ is calibrated from the first task.

\textbf{Metrics.}
Let $A_{t,s}$ denote the accuracy on task $s$ evaluated after training on task $t$. We report the standard continual learning metrics final average accuracy and average forgetting:
\[
\bar{A}_T = \frac{1}{T}\sum_{s=1}^{T} A_{T,s},
\qquad
\bar{F}_T = \frac{1}{T-1}\sum_{s=1}^{T-1}\left(A_{\max,s} - A_{T,s}\right),
\]
where $A_{\max,s} = \max_{t \in \{s,\ldots,T\}} A_{t,s}.$
In the multimodal setting, we additionally report zero-shot transfer on a held-out suite $\mathcal{D}_{\mathrm{zs}}$ consisting of Food101, Oxford-IIIT Pets, Flowers102, Describable Textures (DTD), and Stanford Cars.

\textbf{Compute resources.}
All experiments were run on the Booster partition of CINECA's LEONARDO supercomputer, where each node has four NVIDIA A100 GPUs, each with 64~GB of memory. Each training job used one A100 GPU and 8 CPU cores for data loading under the training schedules described above. The main tables report trainable parameter counts, which characterize the model-side compute scale of the PEFT runs.

\section{Local Score-Stability Diagnostic}
\label{app:local_score_stability}

Section~\ref{sec:local_score_margin} uses projected margin erosion to connect the drift bound in Theorem~\ref{thm:margin_erosion} to forgetting. Here we report the corresponding score-change diagnostic used to estimate the local sensitivity. Theorem~\ref{thm:margin_erosion} bounds old-task margin erosion in terms of parameter drift around the inherited solution. We therefore examine the part of each post-switch trajectory that lies inside the drift budget. For each probed boundary $t\in\{10,50,100\}$, we collect the training steps satisfying $\|\theta_i-\theta_{t-1}^{\star}\|\le\tau$.
On these steps, we measure old-task score changes using the same classwise infinity norm as Assumption~\ref{ass:score_stability}. For each validation example $x$ from old task $q<t$, we compute the score change over all classes seen before task $t$,
\[
\delta_i^{\mathrm{score}}(x)
=
\|s_{1:t-1}(x;\theta_i)-s_{1:t-1}(x;\theta_{t-1}^{\star})\|_\infty .
\]
For each old task, we first compute the maximum of $\delta_i^{\mathrm{score}}(x)$ over its validation examples, and then take the maximum across all old tasks. We denote the resulting worst-case score change by $\Delta_i^{\mathrm{score}}$.

We also measure worst-case projected margin erosion relative to the inherited solution using the same old-class set. Using the margin-erosion definition in Eq.~\eqref{eq:margin_erosion_def}, projected margin erosion is
\[
\Delta\gamma_{t-1}^+(x,y;\theta_i)
=
\max\{0,\Delta\gamma_{t-1}(x,y;\theta_i)\},
\]
where the max operator ensures that margin improvements do not cancel margin losses.

Let $\rho_i=\|\theta_i-\theta_{t-1}^{\star}\|$. For each boundary, we estimate the empirical local sensitivity using probes measured at the end of each training epoch,
\[
\hat C_s=\max_{i\in\mathcal I_{\mathrm{end}}} \frac{\Delta_i^{\mathrm{score}}}{\rho_i},
\]
where $\mathcal I_{\mathrm{end}}$ denotes the end-of-epoch probes among the collected training steps. We then evaluate this bound on held-out probes measured at intermediate training steps within the same epochs. The line $y=\hat C_s x$ gives the estimated worst-case score-stability bound, and $y=2\hat C_s x$ gives the corresponding margin-erosion bound from Theorem~\ref{thm:margin_erosion}.

\begin{figure}[t]
    \centering
    \includegraphics[width=\linewidth]{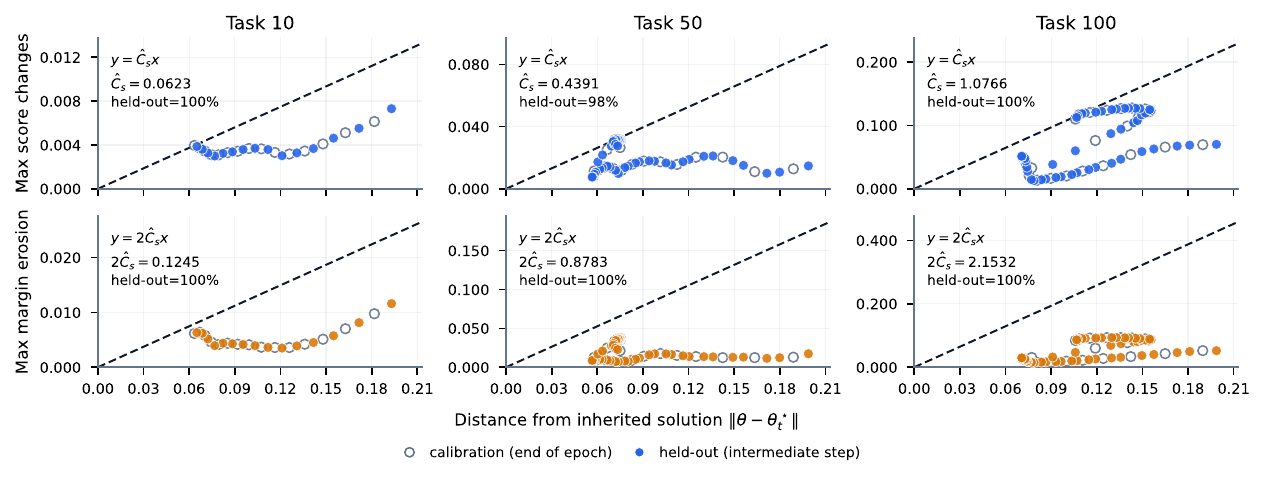}
    \caption{Local score and margin stability inside the drift budget. Top: worst-case old-task score changes versus distance from the inherited solution. Bottom: worst-case projected margin erosion versus distance. Dashed bounds are estimated from end-of-epoch probes and evaluated on held-out intermediate-step probes.}
    \label{fig:local_score_stability_appendix}
\end{figure}

Figure~\ref{fig:local_score_stability_appendix} shows that, after the controller returns the trajectory inside the drift budget, old-task score changes remain well-controlled relative to distance from the inherited solution. The worst-case projected margin erosion is also contained within the corresponding $2\hat C_s$ bound. This supports the local mechanism used by STABLE: by limiting drift around $\theta_{t-1}^{\star}$, the optimizer keeps old-task scores in a regime where inherited margins erode gradually rather than abruptly.

\section{Prediction Preservation Under Controlled Margin Erosion}
\label{app:margin_preservation}

\textbf{Full proof of Theorem~\ref{thm:margin_erosion}.}\label{app:margin_erosion_proof}

\textbf{Restatement of Theorem~\ref{thm:margin_erosion}.}
Assume the Local score stability Assumption~\ref{ass:score_stability} holds with neighborhood $\mathcal U_t$ and constant $C_s$. If $(x,y)$ is a labeled example with $y\in\mathcal Y_{\le t}$ and $\theta\in\mathcal U_t$, then
\[
\gamma_t(x,y;\theta_t^\star)-\gamma_t(x,y;\theta)\le 2C_s\|\theta-\theta_t^\star\|.
\]

\textbf{Proof.}
By Assumption~\ref{ass:score_stability}, for every class $c\in\mathcal Y_{\le t}$,
\[
\left|s_{t,c}(x;\theta)-s_{t,c}(x;\theta_t^\star)\right| \le C_s\|\theta-\theta_t^\star\|.
\]

In particular, this implies
\[
s_{t,y}(x;\theta_t^\star)-s_{t,y}(x;\theta) \le C_s\|\theta-\theta_t^\star\|,
\]
and, for every competing class $c \ne y$,
\[
s_{t,c}(x;\theta)-s_{t,c}(x;\theta_t^\star) \le C_s\|\theta-\theta_t^\star\|.
\]

By definition of the task-$t$ margin,
\[
\gamma_t(x,y;\theta_t^\star)-\gamma_t(x,y;\theta)
=
\bigl[s_{t,y}(x;\theta_t^\star)-\max_{c \ne y} s_{t,c}(x;\theta_t^\star)\bigr]
-
\bigl[s_{t,y}(x;\theta)-\max_{c \ne y} s_{t,c}(x;\theta)\bigr].
\]

Rearranging gives
\[
\gamma_t(x,y;\theta_t^\star)-\gamma_t(x,y;\theta)
=
\bigl[s_{t,y}(x;\theta_t^\star)-s_{t,y}(x;\theta)\bigr]
+
\bigl[\max_{c \ne y} s_{t,c}(x;\theta)-\max_{c \ne y} s_{t,c}(x;\theta_t^\star)\bigr].
\]

For the second term, we use the elementary inequality
\[
\max_i a_i - \max_i b_i \le \max_i(a_i-b_i).
\]

Applying this with $a_c = s_{t,c}(x;\theta)$ and $b_c = s_{t,c}(x;\theta_t^\star)$ over $c \ne y$, we obtain
\[
\max_{c \ne y} s_{t,c}(x;\theta)-\max_{c \ne y} s_{t,c}(x;\theta_t^\star)
\le
\max_{c \ne y} \bigl(s_{t,c}(x;\theta)-s_{t,c}(x;\theta_t^\star)\bigr)
\le
C_s\|\theta-\theta_t^\star\|.
\]

Combining this bound with the bound on the true-class score yields
\[
\gamma_t(x,y;\theta_t^\star)-\gamma_t(x,y;\theta)
\le C_s\|\theta-\theta_t^\star\| + C_s\|\theta-\theta_t^\star\|
= 2C_s\|\theta-\theta_t^\star\|.
\]

This proves the claimed margin erosion bound.

Finally, if
\[
\gamma_t(x,y;\theta_t^\star)>2C_s\|\theta-\theta_t^\star\|,
\]
then the above inequality implies
\[
\gamma_t(x,y;\theta)>0.
\]
Hence the true class $y$ still has strictly larger score than every competing class, so the task-$t$ prediction on $x$ is preserved at $\theta$. \hfill$\square$

\begin{corollary}[Prediction preservation]
\label{cor:margin_preservation}
Any labeled example $(x,y)$ with $y\in\mathcal Y_{\le t}$ satisfying
\[
\gamma_t(x,y;\theta_t^\star)>2C_s\|\theta-\theta_t^\star\|
\]
retains a positive task-$t$ margin, and therefore cannot change its task-$t$ prediction.
\end{corollary}

\textbf{Proof sketch.}
The proof follows directly from Theorem~\ref{thm:margin_erosion}: if $\|\theta-\theta_t^\star\|<\gamma_t(x,y;\theta_t^\star)/(2C_s)$, then $\gamma_t(x,y;\theta)>0$, so the task-$t$ prediction is unchanged.

\section{Adam Moment Reset Baseline}
\label{app:adam_reset}

Table~\ref{tab:adam_reset} tests whether the collapse of Adam is explained only by stale moment estimates carried across task boundaries. Resetting Adam's first- and second-moment estimates at every task switch does not resolve the failure on ImageNet-R with 100 tasks.

\begin{table}[t]
\caption{Adam moment-reset control on ImageNet-R 100-task LoRA. Entries report final average accuracy and forgetting (\%).}
\label{tab:adam_reset}
\centering
\small
\begin{tabular}{lccc}
\toprule
Method & Moment handling & Final avg. acc. & Final avg. forgetting \\
\midrule
Vanilla LoRA + Adam & Carried across tasks & 0.66 & 60.22 \\
Vanilla LoRA + Adam reset & Reset per task & 0.77 & 89.87 \\
Vanilla LoRA + STABLE & Controlled drift & \textbf{73.12} & \textbf{10.17} \\
\bottomrule
\end{tabular}
\end{table}

\section{Fixed L2 Pullback Ablation}
\label{app:fixed_pullback}

We compare STABLE against the most direct static-proximal alternative: vanilla LoRA trained with a fixed L2 pullback to the inherited solution. This baseline uses the same backbone, W-NCM head, data stream, training schedule, and task-wise anchor as STABLE, but replaces the adaptive dual variable with a constant coefficient. It therefore tests whether the gains come merely from anchoring each task to the inherited solution, or from the time-dependent budget feedback used by STABLE. The fixed-pullback baseline trains task $t>1$ with
\[
\mathcal L_t(\theta) + c\|\theta-\theta_{t-1}^{\star}\|_2^2,
\]
where $c$ is a constant coefficient. Table~\ref{tab:fixed_pullback_sweep} sweeps $c$ on ImageNet-A-100 with 50 tasks. Even with the best tuned coefficient, $c=1$, the fixed-pullback baseline reaches only 56.38\% final average accuracy, below STABLE's 62.34\% under the same setting.

In addition, Table~\ref{tab:fixed_pullback_transfer} applies the best coefficient tuned on ImageNet-A-100, $c=1$, to ImageNet-R with 100 tasks. The fixed-pullback baseline performs poorly on this stream, indicating that its limitation is not simply lack of tuning, but the brittleness of a static coefficient under stream-specific and time-varying boundary dynamics.

\begin{table}[t]
\caption{Fixed L2 pullback sweep on ImageNet-A-100 (50 tasks). Entries report accuracy and forgetting (\%) for static coefficients $c$.}
\label{tab:fixed_pullback_sweep}
\centering
\small
\begin{tabular}{ccccc}
\toprule
$c$ & Final avg. acc. & Avg. acc. over tasks & Final avg. forgetting & Max forgetting \\
\midrule
0.01 & 1.29 & 7.47 & 59.19 & 100.00 \\
0.1 & 36.12 & 45.37 & 55.91 & 100.00 \\
0.3 & 42.66 & 61.60 & 41.73 & 90.00 \\
1 & \textbf{56.38} & \textbf{63.40} & 11.09 & \textbf{60.00} \\
3 & 51.93 & 59.06 & 11.26 & 58.33 \\
10 & 52.18 & 60.80 & \textbf{11.06} & \textbf{50.00} \\
\bottomrule
\end{tabular}
\end{table}

\begin{table}[t]
\caption{Fixed L2 pullback transfer test. Entries report final average accuracy (\%) on ImageNet-A-100 and ImageNet-R.}
\label{tab:fixed_pullback_transfer}
\centering
\small
\begin{tabular}{lccc}
\toprule
Method & Coefficient / rule & ImageNet-A-100 & ImageNet-R \\
& & 50 tasks & 100 tasks \\
\midrule
Fixed pullback & $c=1$(best sweep) & 56.38 & 38.84 \\
STABLE & Task-1 budget calibration & \textbf{62.34} & \textbf{73.12} \\
\bottomrule
\end{tabular}
\end{table}

\section{Hard Projection Baseline}
\label{app:hard_projection}

We also compare against a hard-projection variant of Adam. After each Adam update on task $t>1$, we project the LoRA parameters back to the ball around the inherited solution:
\[
\theta_{k+1}
\leftarrow
\theta_{t-1}^{\star}
+
\min\left\{
1,
\frac{\tau}{\|\tilde\theta_{k+1}-\theta_{t-1}^{\star}\|}
\right\}
(\tilde\theta_{k+1}-\theta_{t-1}^{\star}),
\]
where $\tilde\theta_{k+1}$ denotes the unconstrained Adam update. This baseline uses the same backbone, LoRA module, W-NCM head, batch size, learning rate, and task stream as STABLE, but removes the dual variable and replaces the soft adaptive pullback with an explicit projection.

Table~\ref{tab:hard_projection_sweep} sweeps the projection budget on ImageNet-A-100 with 50 tasks. Hard projection can prevent unbounded drift when the budget is tight, but it imposes a fixed constraint throughout training. Small budgets restrict plasticity, while large budgets become too weak to stabilize task-boundary drift. Across the sweep, the best hard-projection result remains below STABLE, supporting the need for an adaptive controller rather than a fixed projection radius.

\begin{table}[t]
\caption{Hard-projection baseline on ImageNet-A-100 (50 tasks). Entries report accuracy and forgetting (\%).}
\label{tab:hard_projection_sweep}
\centering
\small
\begin{tabular}{lccc}
\toprule
Method & Drift Budget & Final avg. acc. & Final avg. forgetting \\
\midrule
\multirow{8}{*}{Hard projection}
& 0.01 & 53.90 & 14.73 \\
& 0.03 & \underline{54.61} & 21.11 \\
& 0.06 & 22.57 & 57.65 \\
& 0.09 & 21.31 & 62.35 \\
& 0.10 & 14.42 & 71.08 \\
& 0.30 & 28.93 & 61.65 \\
& 0.60 & 13.43 & 79.52 \\
& 0.90 & 4.47 & 87.81 \\
\midrule
STABLE & auto calibrated & \textbf{62.34} & \textbf{12.11} \\
\bottomrule
\end{tabular}
\end{table}

\section{Combinations with Replay and Regularization}
\label{app:combined}
Table~\ref{tab:combined} reports additional ImageNet-R experiments with EWC, LwF, and replay. Trained with Adam, each method falls below STABLE alone, and replacing Adam with STABLE improves each of them, most on the 100-task stream (EWC from 0.62\% to 71.57\%, LwF from 0.44\% to 66.24\%, replay from 62.94\% to 75.79\%). EWC and LwF add their own penalties to the drift constraint: with STABLE, they match STABLE alone in accuracy on the 10-task stream but fall below it on the 100-task stream. Replay adds old-task data instead of another penalty and, with STABLE, gives the best accuracy on both streams, a modest gain over STABLE alone. These results show that STABLE already provides strong performance as a standalone optimizer-level method, without requiring standard regularization or replay.

\begin{table}[t]
\caption{STABLE with regularization and replay on ImageNet-R (vanilla LoRA). Entries report accuracy and forgetting (\%) for 10- and 100-task streams. \textbf{Bold}: best per column.}
\label{tab:combined}
\centering
\small
\begin{tabular}{llcccc}
\toprule
& & \multicolumn{2}{c}{10 tasks} & \multicolumn{2}{c}{100 tasks} \\
\cmidrule(lr){3-4} \cmidrule(lr){5-6}
Method & Optimizer & $\bar{A}_T$ & $\bar{F}_T$ & $\bar{A}_T$ & $\bar{F}_T$ \\
\midrule
None & STABLE & 75.74 & 6.09 & 73.12 & 10.17 \\
\midrule
EWC & Adam & 59.95 & 29.30 & 0.62 & 58.47 \\
 & STABLE & 75.95 & 6.67 & 71.57 & 13.58 \\
\midrule
LwF & Adam & 71.74 & 13.57 & 0.44 & 67.29 \\
 & STABLE & 75.76 & \textbf{4.35} & 66.24 & 16.46 \\
\midrule
Replay (2000) & Adam & 73.55 & 12.35 & 62.94 & 25.69 \\
 & STABLE & \textbf{76.84} & 5.28 & \textbf{75.79} & \textbf{7.74} \\
\bottomrule
\end{tabular}
\end{table}

\section{Full results}
\label{app:full_results}

\begin{table*}[!ht]
\caption{Long-horizon unimodal continual learning corresponding to Table~\ref{tab:unimodal_unified}. Metric entries are mean $\pm$ std over 3 seeds for $\bar A_T$ and $\bar F_T$. Joint rows are offline upper bounds; best in \textbf{bold}, second \underline{underlined}.}
\label{tab:full_results}
\centering
\small
{\setlength{\tabcolsep}{3pt}
\resizebox{\textwidth}{!}{%
\begin{tabular}{@{}llcccccc@{}}
\toprule
Method & Opt. & \multicolumn{2}{c}{ImageNet-R (100 tasks)} & \multicolumn{2}{c}{CIFAR-100 (50 tasks)} & \multicolumn{2}{c}{ImageNet-A-100 (50 tasks)} \\
\\
\cmidrule(lr){3-4} \cmidrule(lr){5-6} \cmidrule(l){7-8}
& & $\bar{A}_T$ (\%) & $\bar{F}_T$ (\%) & $\bar{A}_T$ (\%) & $\bar{F}_T$ (\%) & $\bar{A}_T$ (\%) & $\bar{F}_T$ (\%) \\
\midrule
Joint LoRA & Adam & $79.85 \pm 0.20$ & -- & $92.32 \pm 0.18$ & -- & $72.68 \pm 0.39$ & -- \\
Joint Prompts & Adam & $72.23 \pm 0.67$ & -- & $88.86 \pm 0.15$ & -- & $61.34 \pm 0.52$ & -- \\
\midrule
\multicolumn{8}{l}{\textit{LoRA-based methods}} \\
O-LoRA~\cite{Wang2023OLoRA} & AdamW & $57.87 \pm 2.62$ & $33.87 \pm 2.41$ & $74.09 \pm 1.34$ & $25.90 \pm 1.39$ & $26.92 \pm 2.21$ & $59.37 \pm 1.56$ \\
HiDeLoRA~\cite{Wang2025HiDePET} & Adam & $72.36 \pm 0.62$ & \underline{6.16 $\pm$ 0.44} & $79.92 \pm 2.70$ & \textbf{2.29 $\pm$ 0.06} & $60.37 \pm 0.97$ & $10.62 \pm 1.20$ \\
InfLoRA~\cite{Liang2024} & Adam & $72.54 \pm 1.56$ & $18.05 \pm 2.35$ & $84.93 \pm 3.63$ & $13.74 \pm 3.99$ & \underline{61.69 $\pm$ 0.76} & $22.33 \pm 0.45$ \\
Vanilla LoRA & Adam & $0.66 \pm 0.14$ & $60.22 \pm 2.12$ & $1.30 \pm 0.06$ & $75.70 \pm 4.68$ & $1.25 \pm 0.06$ & $58.41 \pm 0.39$ \\
Vanilla LoRA & \textbf{STABLE} & \textbf{73.12 $\pm$ 0.43} & $10.17 \pm 0.06$ & \textbf{88.77 $\pm$ 0.23} & $7.90 \pm 0.24$ & \textbf{62.34 $\pm$ 1.30} & $12.11 \pm 0.80$ \\
\midrule
\multicolumn{8}{l}{\textit{Prompt-based methods}} \\
L2P~\cite{Wang2022} & Adam & $68.60 \pm 0.23$ & $8.74 \pm 0.30$ & $85.32 \pm 0.31$ & $8.31 \pm 0.41$ & $19.13 \pm 0.92$ & \underline{9.07 $\pm$ 1.38} \\
DualPrompt~\cite{Wang2022dual} & Adam & $70.39 \pm 0.71$ & $10.69 \pm 0.46$ & $86.25 \pm 0.95$ & $9.21 \pm 1.10$ & $19.37 \pm 0.34$ & \textbf{7.62 $\pm$ 0.45} \\
CODA-Prompt~\cite{Smith2023CODAPrompt} & Adam & $69.32 \pm 0.16$ & \textbf{6.07 $\pm$ 0.26} & $87.72 \pm 0.15$ & $5.35 \pm 0.27$ & $59.37 \pm 0.87$ & $10.67 \pm 0.84$ \\
Vanilla Prompts & Adam & $68.01 \pm 0.48$ & $16.15 \pm 0.74$ & $83.88 \pm 0.83$ & $12.00 \pm 1.19$ & $56.31 \pm 0.95$ & $22.13 \pm 1.31$ \\
Vanilla Prompts & \textbf{STABLE} & \underline{72.99 $\pm$ 0.26} & $7.64 \pm 0.36$ & \underline{87.88 $\pm$ 0.07} & \underline{4.99 $\pm$ 0.19} & $58.88 \pm 0.40$ & $14.23 \pm 0.50$ \\
\bottomrule
\end{tabular}
}
}
\end{table*}

\begin{table*}[!ht]
\caption{CLIP multimodal continual learning corresponding to Table~\ref{tab:multimodal_unified}. Metric entries are mean $\pm$ std over 3 seeds for $\bar A_T$ and $\bar F_T$. The joint row is an offline upper bound; best in \textbf{bold}, second \underline{underlined}.}
\label{tab:multimodal_unified_full}
\centering
\small
{\setlength{\tabcolsep}{2pt}
\renewcommand{\arraystretch}{1.12}
\resizebox{\textwidth}{!}{%
\begin{tabular}{@{}llcccccc@{}}
\toprule
Method & Opt. & \multicolumn{2}{c}{ImageNet-R} & \multicolumn{2}{c}{CIFAR-100} & \multicolumn{2}{c}{ImageNet-A-100} \\
\cmidrule(lr){3-4} \cmidrule(lr){5-6} \cmidrule(l){7-8}
& & $\bar{A}_T$ (\%) & $\bar{F}_T$ (\%) & $\bar{A}_T$ (\%) & $\bar{F}_T$ (\%) & $\bar{A}_T$ (\%) & $\bar{F}_T$ (\%) \\
\midrule
Joint LoRA & Adam & $86.06 \pm 0.17$ & -- & $87.45 \pm 0.11$ & -- & $74.28 \pm 0.51$ & -- \\
\midrule
Vanilla LoRA & Adam & $64.34 \pm 1.68$ & $25.74 \pm 2.10$ & $56.84 \pm 1.47$ & $44.68 \pm 1.74$ & $28.15 \pm 1.18$ & $55.03 \pm 0.38$ \\
\midrule
MindTheGap~\cite{Huang2025ICCV} & Adam & \textbf{81.50 $\pm$ 0.08} & $6.94 \pm 0.14$ & $80.40 \pm 0.10$ & $9.62 \pm 0.15$ & $65.98 \pm 0.78$ & $6.59 \pm 0.21$ \\
RAPF~\cite{Huang2024RAPF} & Adam & $80.41 \pm 0.12$ & \underline{3.30 $\pm$ 0.26} & \underline{80.41 $\pm$ 0.11} & \textbf{6.99 $\pm$ 0.11} & $61.07 \pm 1.06$ & $16.57 \pm 1.97$ \\
MoE4CL~\cite{Yu2024MoE4CL} & AdamW & $80.37 \pm 0.18$ & $5.48 \pm 0.08$ & $79.08 \pm 0.06$ & $9.64 \pm 0.09$ & $58.20 \pm 0.14$ & \textbf{5.75 $\pm$ 0.17} \\
CLAP4CLIP~\cite{Jha2024CLAP4CLIP} & SGD & $79.37 \pm 0.48$ & $7.90 \pm 0.47$ & $78.61 \pm 0.13$ & $9.92 \pm 0.09$ & \underline{67.27 $\pm$ 0.37} & $11.69 \pm 0.70$ \\
ENGINE~\cite{Zhou2025ENGINE} & SGD & $76.80 \pm 0.06$ & $4.86 \pm 0.11$ & $73.85 \pm 0.08$ & $12.62 \pm 0.22$ & $57.22 \pm 0.15$ & $11.52 \pm 0.49$ \\
Vanilla LoRA & \textbf{STABLE} & \underline{81.28 $\pm$ 0.14} & \textbf{2.80 $\pm$ 0.20} & \textbf{81.61 $\pm$ 0.16} & \underline{9.10 $\pm$ 0.07} & \textbf{70.19 $\pm$ 0.37} & \underline{6.54 $\pm$ 0.17} \\
\bottomrule
\end{tabular}
}
}
\end{table*}

\begin{table}[t]
\caption{CLIP zero-shot accuracy (\%) after the ImageNet-R stream. Entries are mean $\pm$ std over 3 seeds using the standard image-text head. Pre-trained CLIP and joint rows are references; best in \textbf{bold}, second \underline{underlined}.}
\label{tab:multimodal_zeroshot_full_imagenet_r}
\centering
\small
{\setlength{\tabcolsep}{3pt}
\renewcommand{\arraystretch}{1.12}
\begin{tabular}{@{}lccccc@{}}
\toprule
Method & Food101 & Pets & Flowers & DTD & Cars \\
\midrule
Pre-trained CLIP & 88.38 & 88.01 & 66.03 & 44.31 & 64.42 \\
\midrule
Joint LoRA + Adam & $88.32 \pm 0.06$ & $89.76 \pm 0.33$ & $68.35 \pm 0.23$ & $44.75 \pm 0.31$ & $63.10 \pm 0.07$ \\
\midrule
Vanilla LoRA + Adam & $74.76 \pm 1.32$ & $75.80 \pm 2.06$ & $52.51 \pm 0.07$ & $38.67 \pm 0.70$ & $48.90 \pm 0.95$ \\
\midrule
MindTheGap & $85.97 \pm 0.55$ & \underline{87.40 $\pm$ 0.40} & \underline{65.43 $\pm$ 2.21} & \underline{43.78 $\pm$ 0.44} & \underline{60.32 $\pm$ 1.19} \\
RAPF & $47.01 \pm 2.35$ & $69.91 \pm 0.74$ & $32.06 \pm 4.32$ & $24.77 \pm 0.97$ & $23.90 \pm 2.55$ \\
MoE4CL & $85.12 \pm 0.21$ & $84.45 \pm 0.13$ & $61.66 \pm 0.31$ & $43.62 \pm 0.44$ & $59.69 \pm 0.59$ \\
CLAP4CLIP & \underline{86.20 $\pm$ 0.84} & $80.34 \pm 4.09$ & $58.91 \pm 4.76$ & $40.60 \pm 2.06$ & $59.51 \pm 1.56$ \\
ENGINE & $37.57 \pm 0.27$ & $59.05 \pm 0.32$ & $22.41 \pm 0.35$ & $20.94 \pm 0.13$ & $17.54 \pm 0.15$ \\
Vanilla LoRA + STABLE & \textbf{88.56 $\pm$ 0.14} & \textbf{88.77 $\pm$ 0.18} & \textbf{67.86 $\pm$ 0.41} & \textbf{46.05 $\pm$ 0.13} & \textbf{64.19 $\pm$ 0.25} \\
\bottomrule
\end{tabular}
}
\end{table}

\begin{table}[t]
\caption{CLIP zero-shot accuracy (\%) after the CIFAR-100 stream. Entries are mean $\pm$ std over 3 seeds using the standard image-text head. Pre-trained CLIP and joint rows are references; best in \textbf{bold}, second \underline{underlined}.}
\label{tab:multimodal_zeroshot_full_cifar100}
\centering
\small
{\setlength{\tabcolsep}{3pt}
\renewcommand{\arraystretch}{1.12}
\begin{tabular}{@{}lccccc@{}}
\toprule
Method & Food101 & Pets & Flowers & DTD & Cars \\
\midrule
Pre-trained CLIP & 88.38 & 88.01 & 66.03 & 44.31 & 64.42 \\
\midrule
Joint LoRA + Adam & $86.35 \pm 0.07$ & $88.16 \pm 0.54$ & $68.68 \pm 0.56$ & $44.29 \pm 0.36$ & $61.61 \pm 0.33$ \\
\midrule
Vanilla LoRA + Adam & $58.68 \pm 3.01$ & $62.66 \pm 3.18$ & $33.85 \pm 0.66$ & $32.94 \pm 1.11$ & $38.20 \pm 3.98$ \\
\midrule
MindTheGap & \underline{87.85 $\pm$ 0.31} & \underline{86.21 $\pm$ 1.13} & \textbf{68.81 $\pm$ 0.38} & \underline{44.72 $\pm$ 0.13} & \underline{61.73 $\pm$ 0.31} \\
RAPF & $18.71 \pm 0.64$ & $26.02 \pm 5.15$ & $10.73 \pm 1.15$ & $15.05 \pm 1.05$ & $6.17 \pm 0.37$ \\
MoE4CL & $86.35 \pm 0.53$ & $80.43 \pm 0.21$ & $66.88 \pm 1.43$ & $41.35 \pm 0.62$ & $55.48 \pm 0.75$ \\
CLAP4CLIP & $85.94 \pm 0.30$ & $80.65 \pm 2.39$ & $60.25 \pm 2.94$ & $41.22 \pm 1.01$ & $60.17 \pm 0.99$ \\
ENGINE & $20.46 \pm 0.41$ & $23.26 \pm 0.41$ & $16.98 \pm 0.19$ & $13.12 \pm 0.07$ & $9.15 \pm 0.19$ \\
Vanilla LoRA + STABLE & \textbf{88.63 $\pm$ 0.03} & \textbf{88.62 $\pm$ 0.05} & \underline{67.73 $\pm$ 0.10} & \textbf{44.84 $\pm$ 0.11} & \textbf{64.11 $\pm$ 0.08} \\
\bottomrule
\end{tabular}
}
\end{table}

\clearpage
\begin{table}[t]
\caption{CLIP zero-shot accuracy (\%) after the ImageNet-A-100 stream. Entries are mean $\pm$ std over 3 seeds using the standard image-text head. Pre-trained CLIP and joint rows are references; best in \textbf{bold}, second \underline{underlined}.}
\label{tab:multimodal_zeroshot_full_imagenet_a}
\centering
\small
{\setlength{\tabcolsep}{3pt}
\renewcommand{\arraystretch}{1.12}
\begin{tabular}{@{}lccccc@{}}
\toprule
Method & Food101 & Pets & Flowers & DTD & Cars \\
\midrule
Pre-trained CLIP & 88.38 & 88.01 & 66.03 & 44.31 & 64.42 \\
\midrule
Joint LoRA + Adam & $87.70 \pm 0.06$ & $87.94 \pm 0.62$ & $67.86 \pm 0.54$ & $44.04 \pm 0.23$ & $62.58 \pm 0.36$ \\
\midrule
Vanilla LoRA + Adam & $36.58 \pm 4.31$ & $40.86 \pm 2.83$ & $17.01 \pm 5.58$ & $23.40 \pm 1.33$ & $26.86 \pm 2.82$ \\
\midrule
MindTheGap & \underline{88.17 $\pm$ 0.03} & \underline{87.16 $\pm$ 0.36} & \underline{65.84 $\pm$ 0.37} & $43.62 \pm 0.05$ & \textbf{63.54 $\pm$ 0.09} \\
RAPF & $8.77 \pm 2.61$ & $13.89 \pm 2.95$ & $7.67 \pm 0.63$ & $9.72 \pm 0.46$ & $5.20 \pm 1.09$ \\
MoE4CL & $87.98 \pm 0.02$ & \textbf{87.43 $\pm$ 0.05} & \textbf{66.12 $\pm$ 0.03} & \underline{43.87 $\pm$ 0.09} & 63.42 $\pm$ 0.07 \\
CLAP4CLIP & $86.81 \pm 0.57$ & $84.82 \pm 1.80$ & $64.35 \pm 1.10$ & $41.22 \pm 0.49$ & $62.56 \pm 0.08$ \\
ENGINE & $8.56 \pm 0.51$ & $11.32 \pm 0.91$ & $8.94 \pm 1.72$ & $9.36 \pm 0.33$ & $1.94 \pm 0.31$ \\
Vanilla LoRA + STABLE & \textbf{88.25 $\pm$ 0.16} & $87.04 \pm 0.21$ & $65.65 \pm 1.04$ & \textbf{44.08 $\pm$ 0.22} & \underline{$63.51 \pm 0.19$} \\
\bottomrule
\end{tabular}
}
\end{table}

\begin{table*}[t]
\caption{Common-head CLIP multimodal continual learning. Metric entries are mean $\pm$ std over 3 seeds for $\bar A_T$ and $\bar F_T$, with all methods evaluated by W-NCM. The joint row is an offline upper bound; best in \textbf{bold}, second \underline{underlined}.}
\label{tab:multimodal_shared_wncm_full}
\centering
\small
{\setlength{\tabcolsep}{2pt}
\renewcommand{\arraystretch}{1.12}
\resizebox{\textwidth}{!}{%
\begin{tabular}{@{}llcccccc@{}}
\toprule
Method & Opt. & \multicolumn{2}{c}{ImageNet-R} & \multicolumn{2}{c}{CIFAR-100} & \multicolumn{2}{c}{ImageNet-A-100} \\
\cmidrule(lr){3-4} \cmidrule(lr){5-6} \cmidrule(l){7-8}
& & $\bar{A}_T$ (\%) & $\bar{F}_T$ (\%) & $\bar{A}_T$ (\%) & $\bar{F}_T$ (\%) & $\bar{A}_T$ (\%) & $\bar{F}_T$ (\%) \\
\midrule
Joint LoRA & Adam & $83.71 \pm 0.16$ & -- & $87.43 \pm 0.24$ & -- & $73.00 \pm 0.18$ & -- \\
\midrule
Vanilla LoRA & Adam & $64.34 \pm 1.68$ & $25.74 \pm 2.10$ & $56.84 \pm 1.47$ & $44.68 \pm 1.74$ & $28.15 \pm 1.18$ & $55.03 \pm 0.38$ \\
\midrule
MindTheGap~\cite{Huang2025ICCV} & Adam & $79.63 \pm 0.32$ & $8.46 \pm 0.49$ & $79.76 \pm 0.46$ & $15.76 \pm 0.44$ & \underline{69.63 $\pm$ 0.19} & $7.07 \pm 0.69$ \\
RAPF~\cite{Huang2024RAPF} & Adam & $48.09 \pm 0.49$ & $42.36 \pm 0.48$ & $48.46 \pm 0.26$ & $51.29 \pm 0.14$ & $34.64 \pm 0.91$ & $52.54 \pm 1.17$ \\
MoE4CL~\cite{Yu2024MoE4CL} & AdamW & \underline{80.21 $\pm$ 0.16} & $6.41 \pm 0.11$ & \underline{79.85 $\pm$ 0.10} & $15.89 \pm 0.14$ & $68.23 \pm 0.08$ & \textbf{6.44 $\pm$ 0.09} \\
CLAP4CLIP~\cite{Jha2024CLAP4CLIP} & SGD & $77.56 \pm 0.05$ & \underline{5.61 $\pm$ 0.09} & $78.31 \pm 0.02$ & \textbf{8.95 $\pm$ 0.08} & $63.47 \pm 0.41$ & $12.52 \pm 0.86$ \\
ENGINE~\cite{Zhou2025ENGINE} & SGD & $54.20 \pm 0.34$ & $33.59 \pm 0.46$ & $45.08 \pm 0.07$ & $53.46 \pm 0.10$ & $27.69 \pm 0.49$ & $53.94 \pm 0.57$ \\
Vanilla LoRA & \textbf{STABLE} & \textbf{81.28 $\pm$ 0.14} & \textbf{2.80 $\pm$ 0.20} & \textbf{81.61 $\pm$ 0.16} & \underline{9.10 $\pm$ 0.07} & \textbf{70.19 $\pm$ 0.37} & \underline{6.54 $\pm$ 0.17} \\
\bottomrule
\end{tabular}
}
}
\end{table*}

\section{Sensitivity to the Dual Step Size}
\label{app:eta_sensitivity}

The dual update of Eq.~\eqref{eq:duco_dual_update} has a dual step size $\eta$, set to 0.03 in all other experiments. To test its sensitivity, we rerun ImageNet-R with 100 tasks, keeping the configuration of Section~\ref{sec:experiments}, including its primal learning rate of $5\times10^{-4}$, and varying only $\eta$. Across a 60-fold range of $\eta$, accuracy stays within 1.4 points and forgetting within 1.1 points (Table~\ref{tab:eta_sensitivity}). The two other controller constants, $\lambda_{\mathrm{init}}$ and $\epsilon_{\mathrm{dz}}$, are the same in every experiment: the initial dual value sets $\lambda$ only at the start of Task~2, after which the dual update adjusts it, and the deadzone is a noise tolerance that avoids oscillation at the constraint boundary.

\begin{table}[ht]
\caption{Sensitivity to the dual step size $\eta$ on ImageNet-R (100 tasks, vanilla LoRA). Entries report final average accuracy and forgetting (\%).}
\label{tab:eta_sensitivity}
\centering
\small
\begin{tabular}{lcccccc}
\toprule
$\eta$ & 0.01 & 0.03 (default) & 0.06 & 0.1 & 0.3 & 0.6 \\
\midrule
Final avg. acc. & 72.62 & 73.12 & 73.61 & 73.18 & 73.96 & 73.92 \\
Avg. forgetting & 10.48 & 10.17 & 9.39 & 10.27 & 9.72 & 9.64 \\
\bottomrule
\end{tabular}
\end{table}

\section{Linear Classifier Heads}
\label{app:linear_head}

We use a distance-based head for two reasons. First, it matches the representation-induced score setting analyzed in Theorem~\ref{thm:margin_erosion}: without separate trainable classifier weights, parameter drift directly induces score and margin changes. Second, it gives the tested baselines higher accuracy. Table~\ref{tab:linear_head} evaluates L2P, DualPrompt, and InfLoRA with the native linear head of their original implementations and with the W-NCM head of Table~\ref{tab:unimodal_unified}. Linear classifiers are known to become biased toward recently learned classes~\cite{Wu2019BiC}, and nearest-class-mean classifiers reduce this recency bias~\cite{Mai2021SCR}. The gap between the two heads is larger on the 100-task stream than on the 50-task stream, consistent with this bias growing with the number of tasks. A linear head would therefore substantially weaken these baselines, so all methods in Table~\ref{tab:unimodal_unified} use the same W-NCM head.

\begin{table}[ht]
\caption{Baselines with their native linear head and with the W-NCM head of Table~\ref{tab:unimodal_unified}. Entries report final average accuracy (\%), mean $\pm$ std over 3 seeds.}
\label{tab:linear_head}
\centering
\small
\begin{tabular}{llcc}
\toprule
Stream & Method & Linear head & W-NCM head \\
\midrule
\multirow{3}{*}{ImageNet-R, 100 tasks} & L2P & $34.43 \pm 0.55$ & $68.60 \pm 0.23$ \\
 & DualPrompt & $27.44 \pm 1.08$ & $70.39 \pm 0.71$ \\
 & InfLoRA & $38.90 \pm 0.14$ & $72.54 \pm 1.56$ \\
\midrule
\multirow{3}{*}{CIFAR-100, 50 tasks} & L2P & $63.72 \pm 1.30$ & $85.32 \pm 0.31$ \\
 & DualPrompt & $68.63 \pm 0.50$ & $86.25 \pm 0.95$ \\
 & InfLoRA & $58.13 \pm 4.90$ & $84.93 \pm 3.63$ \\
\bottomrule
\end{tabular}
\end{table}

\section{Forgetting over Long Streams}
\label{app:long_stream}

STABLE limits the drift from the inherited solution within each task. Our longest experiments test whether this limited per-task drift accumulates into large forgetting over a long stream. Final average accuracy and average forgetting in Table~\ref{tab:unimodal_unified} are measured after the entire stream, so any accumulated erosion of early tasks would appear there. On ImageNet-R, STABLE ends at 73.12\% with 10.17\% forgetting after 100 tasks, and at 75.74\% on the 10-task stream (Table~\ref{tab:combined}), so splitting the same classes into ten times as many tasks costs 2.6 points of accuracy.

In the diagnostic run shown in Figures~\ref{fig:post_switch_controller} and~\ref{fig:local_margin_forgetting}, average forgetting over old tasks is 4.15\% after Task~10, 8.63\% after Task~50, and 9.88\% after Task~100. The last fifty task switches add only 1.25 points, against 4.48 points for the forty switches from Task~10 to Task~50, so forgetting grows more slowly as the stream gets longer. Accumulated drift would also make the oldest tasks the most forgotten, but at the end of the stream the oldest third of tasks shows 11.73\% mean forgetting, no more than the middle third (12.40\%), and retains 72.21\% accuracy.

We attribute this to the anchor refresh: each boundary moves the anchor to a solution that already supports all previous tasks, and training on the next task ends close to it. Figure~\ref{fig:post_switch_controller} shows the trajectory settling well inside the budget after the post-switch transients at Tasks~10, 50, and~100; in the same run, the end-of-task distances from the anchor are 0.08 to 0.10, against a budget of $\tau=0.203$.

\end{document}